\documentclass[a4paper,fleqn]{cas-sc}

\usepackage[authoryear,longnamesfirst]{natbib}
\usepackage{amsmath,amssymb}
\usepackage{graphicx}
\usepackage{booktabs}
\usepackage{multirow}
\usepackage{siunitx}
\usepackage{microtype}
\usepackage{url}
\hypersetup{hidelinks}
\usepackage[nameinlink,noabbrev]{cleveref}
\usepackage{subcaption}

\newcommand{\methodname}{DIP-CAD}

\begin{document}

% Uncomment if line numbering is requested:
% \linenumbers

\shorttitle{DIP-CAD: Design-Intent-Preserving Parametric CAD Modeling}
\shortauthors{X. Cheng et al.}

\title[mode=title]{A Design-Intent-Preserving Agent Harness for Parametric CAD Modeling}

\author[1]{Xi Cheng}
\fnmark[1]
\ead{chengxi23@tsinghua.org.cn}
\author[1]{Chenxi Zhai}
\fnmark[1]
\ead{dcx24@mails.tsinghua.edu.cn}
\author[1]{Jie Xu}
\fnmark[1]
\ead{xuj22.thu@gmail.com}
\author[1]{Hang Cheng}
\ead{chenghan24@mails.tsinghua.edu.cn}
\author[1]{Mingyu Fan}
\ead{my-fan25@mails.tsinghua.edu.cn}
\author[1,2]{Pingfa Feng}
\ead{fengpf@tsinghua.edu.cn}
\author[1]{Long~Zeng}
\cormark[1]
\ead{zenglong@sz.tsinghua.edu.cn}
\cortext[1]{Corresponding author.}
\fntext[1]{These authors contributed equally to this work.}

\affiliation[1]{
  organization={Tsinghua Shenzhen International Graduate School, Tsinghua University},
  % addressline={Street and number},
  city={Shenzhen},
  postcode={518055},
  country={China}
}

\affiliation[2]{
  organization={Department of Mechanical Engineering, Tsinghua University},
  % addressline={Street and number},
  city={Beijing},
  postcode={100084},
  country={China}
}

\begin{abstract}
Parametric CAD modeling requires not only matching the target shape, but also preserving design intent under parameter edits. However, most existing methods focus on improving static shape generation accuracy, while the dependencies that govern later edits are often overlooked.
To tackle this issue, we present a design-intent-preserving agent harness for generating parametric CAD models (\methodname). 
\methodname\ combines vision-language model reasoning with CAD kernel operations to establish dependencies that encode the design intent. In parametric CAD, these dependencies are mainly established through geometry references in modeling commands and constraints in parametric sketches. 
At the command level, ID-based explicit geometry reference binds modeling commands to selected faces, edges, or vertices.
At the sketch level, a local-coordinate bridge ensures accurate sketch placement, providing the basis for correctly establishing sketch constraints. Constraints among sketch primitives and constraints to external projected geometry further preserve both internal sketch relationships and dependencies on external geometry.
Additionally, to provide a more comprehensive data foundation for parametric CAD modeling tasks, we construct the Geometry Explicit Reference Dataset (GERD), whose structured data format links stepwise modeling commands to intermediate CAD models through explicit geometry references. 
On GERD-EVL and a DeepCAD test subset, \methodname\ improves mIoU for static model generation by 11.1\% and 5.6\%, respectively. In parameter-editing evaluations, the corresponding mIoU improvements are 7.0\% and 4.2\%. Case studies further demonstrate how parameter edits drive shape updates while preserving design intent.
\end{abstract}


\maketitle

\section{Introduction}
\label{sec:introduction}

Parametric CAD models encode design intent through geometry reference in modeling commands and constraints in parametric sketches. When upstream parameters are edited, these dependencies enable the CAD models to update while preserving design intent \citep{yin2012featureconstraints,cheng2019designintent}. 
This editability makes parametric CAD models fundamental to iterative design, engineering analysis, and manufacturing. Meanwhile, product images and renderings are widely available during engineering design and product development. Automatically converting such visual inputs into parametric CAD models can reduce the manual effort required for CAD modeling. The resulting models can then be directly edited and reused in engineering tasks, including analysis, optimization, and manufacturing~\citep{chen2025img2cad,chen2025cadcrafter}.

\begin{figure}[pos=htbp]
    \centering
    \includegraphics[width=1.0\linewidth]{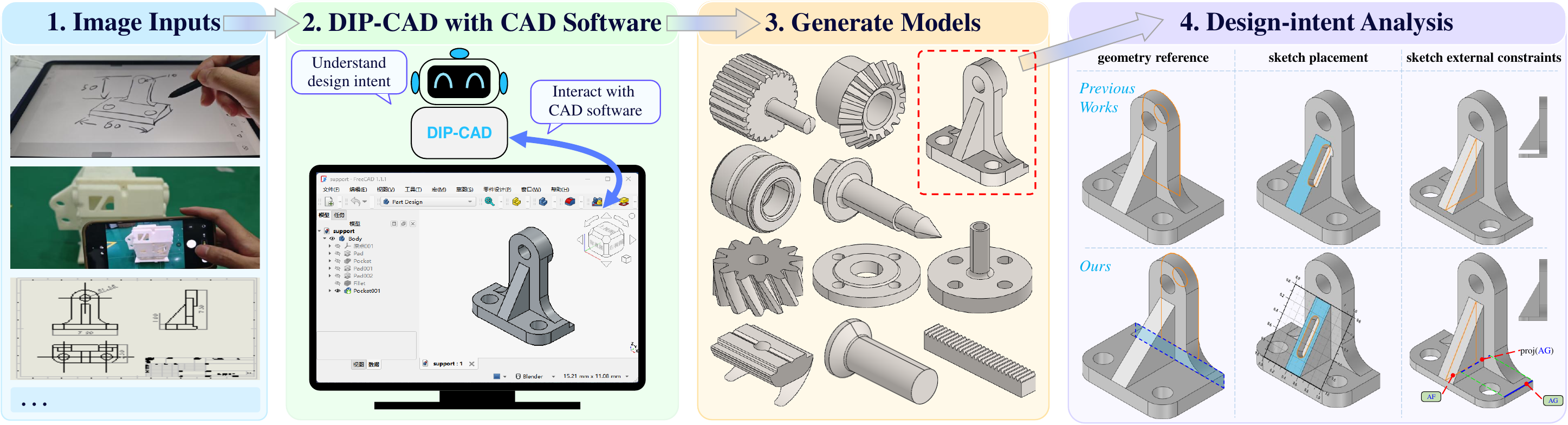}
    \caption{\textbf{\methodname\ overview.} Given a part image, DIP-CAD understands the design intent, and then interacts with CAD software to generate native parametric CAD models. \textbf{4-Left:} referencing the back blue face enables the back plate follow the back face when base width increases. \textbf{4-Middle:} exposing the sketch plane's local coordinate system explicitly ensures correct slot profile placement on the inclined face. \textbf{4-Right:} constraints to projected external geometry keep the rib connected to the base after base height decreases.}
    \label{fig:comp_prev}
\end{figure}

Existing CAD generation methods have improved static model generation accuracy, but preserving the design intents that govern parameter edits remains a challenge. 
Early methods use transformer or diffusion to generate CAD models as sequences of predefined commands. DeepCAD represents sketch primitives, sketch planes, and extrusion parameters as command vectors and learns their sequential structure with a Transformer~\citep{wu2021deepcad}. Text2CAD introduces text-conditioned sequence generation, while CAD-Diffuser generates modeling sequences from point clouds through multimodal diffusion~\citep{khan2024text2cad,ma2024caddiffuser}. Image-conditioned methods, including Img2CAD and CADCrafter, further connect visual inputs to structured CAD generation~\citep{chen2025img2cad,chen2025cadcrafter}.
These methods provide compact representations for learning sketch-and-extrude modeling sequences. However, their command vocabularies primarily describe what type of command to perform and its parameters, without explicitly identifying existing faces, edges, or vertices as references. Consequently, reproducing the target shape does not ensure that downstream commands or sketches follow upstream changes as intended.
Subsequent LLM- and VLM-based methods typically generate executable CAD code, allowing them to use the reference mechanisms of CAD libraries such as CadQuery. CAD-Recode translates point clouds into Python CAD code~\citep{rukhovich2025cadrecode}, and Ortho2CAD generates CadQuery code from drawings~\citep{joglekar2026ortho2cad}. CAD-Coder and ReCAD incorporates reinforcement learning to improve CAD code generation performance \citep{doris2026cadcoder,li2026recad}.
However, these methods primarily generate the complete CAD code without observing intermediate models, making reference selection difficult to verify and correct~\citep{rukhovich2025cadrecode,joglekar2026ortho2cad,doris2026cadcoder}.
Recent agent-based methods incorporate execution, inspection, and revision to generate CAD code or native parametric CAD models (e.g., .SLDPRT document for Solidworks, and .FCStd document for FreeCAD). CAD-Assistant provides FreeCAD tools and visual feedback, while Seek-CAD combines stepwise feedback with references to start, end, and swept faces~\citep{mallis2025cadassistant,li2026seekcad}. Multi-agent systems include CADSmith, which combines error correction, geometric validation, and VLM assessment, and Fusion AI-AD, which integrates design knowledge processing with SolidWorks modeling~\citep{barkley2026cadsmith,liu2025fusionaiad}. These works enable some reference errors to be corrected.
Related methods further incorporate geometric information from intermediate models. CAD-Refiner uses a topology graph and geometry checker~\citep{yuan2026cadrefiner}. Pointer-CAD encodes the current boundary representation (B-Rep) to support soft face and edge selection~\citep{qi2026pointercad}.
Nevertheless, feature-matching-based geometry reference adopted by these methods remains prone to errors when multiple faces or edges exhibit similar features, leading to incorrect reference selection. Such errors can propagate to sketch placement and constraints to projected external geometry, undermining their ability to faithfully preserve design intent (\Cref{fig:comp_prev}).

To address these limitations, we introduce \methodname, a design-intent-preserving agent harness. It enables parametric CAD modeling that mirrors engineering workflows by customizing the interaction between a vision-language model (VLM) and a CAD kernel.
Parametric CAD models are typically constructed through modeling commands and parametric sketches~\citep{camba2016parametric,kyratzi2022integrated,company2020constraints}. For modeling commands, a key challenge is accurately selecting existing geometry. \methodname\ addresses this challenge using edge-ID tags and structured geometry tables for explicit geometry reference.
For parametric sketches, two key challenges arise. The first is accurately placing sketch primitives on the sketch plane, particularly for inclined supports. The second is establishing constraints between sketch primitives and projected external geometry. \methodname\ addresses these challenges through a view-to-local coordinate bridge and automatic constraint generation, respectively (\Cref{fig:comp_prev}).
Additionally, existing CAD datasets often lack step-level explicit geometry reference annotations, and native parametric CAD files are incompatible across CAD systems. We design an explicit geometric reference data format. At each modeling step, this format records both the modeling command and the model after its execution. It assigns explicit IDs to intermediate models, allowing modeling commands to reference existing geometry through these IDs. Based on this format, we construct the Geometry Explicit Reference Dataset (GERD).
On GERD-EVL and a DeepCAD test subset, \methodname\ improves static reconstruction mIoU by 11.1\% and 5.6\%, respectively. To evaluate parameter editability, we increase all extrude depths by 10\% and assess the updated models. \methodname\ achieves mIoU improvements of 7.0\% and 4.2\%, respectively. We further conduct case studies with sequential parameter edits. These cases illustrate how parameters drive shape updates. Together, these results demonstrate the effectiveness of \methodname\ in both static model generation and design intent preservation.

The main contributions of this work are as follows: 
\begin{itemize}
    \item We characterize design intent as the dependencies in modeling commands and parametric sketches, identify three key challenges in constructing these dependencies, and introduce corresponding mechanisms to tackle these challenges.
    
    \item We develop \methodname, a design-intent-preserving agent harness for generating parametric CAD models from images. \methodname\ performs stepwise command generation, sketch generation, and modeling verification to generate native parametric CAD documents (.FCStd files for FreeCAD). 
    
    \item We design an explicit geometry-reference data format and construct the Geometry Explicit Reference Dataset (GERD). GERD contains 2,700 models with step-aligned commands, intermediate geometric states, and geometry IDs. GERD provides the evolving geometric context and explicit entity references required for studying parametric CAD generation.
    
    \item We evaluate both static model generation and parameter edit performance on GERD-EVL and a DeepCAD test subset. \methodname\ achieves the best performance. Parameter editing cases and ablation studies further demonstrate the effectiveness of the proposed mechanisms for preserving design intents.

\end{itemize}

\section{Problem Statement}
\label{sec:challenges}

\begin{figure}[pos=htbp]
    \centering
    \includegraphics[width=0.65\linewidth]{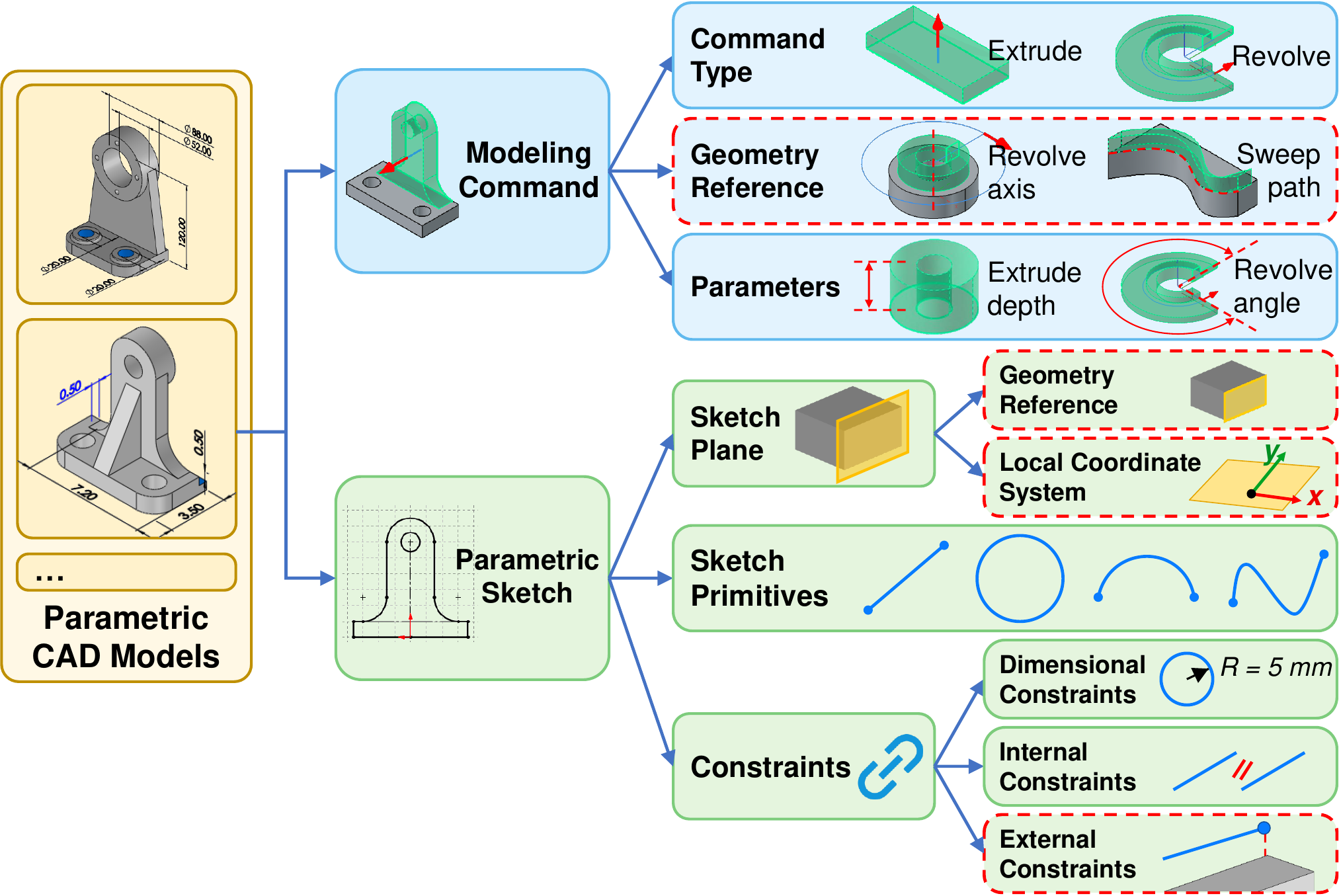}
    \caption{\textbf{Key Elements of Parametric CAD Model.} The three key challenges of generating Parametric CAD Model are Geometry Reference, Sketch Placement (Local coordinate system), and External Constraints.}
    \label{fig:paramcad-ele}
\end{figure}

Parametric CAD models typically consist of modeling commands and parametric sketches, as illustrated in \Cref{fig:paramcad-ele}. This structure highlights why predicting command types, numerical parameters, and sketch primitives alone is insufficient for design-intent-preserving CAD generation.

% Above decomposition explains why predicting command types, parameter values, and sketch primitives is insufficient for design-intent-preserving generation. The decomposition leads to three core challenges: Geometry Reference, Sketch Placement, and Projected External Geometry Constraints.

\subsection{Geometry Reference}
\label{subsec:dependency_reference}

\begin{figure}[pos=htbp]
    \centering
    \includegraphics[width=0.65\linewidth]{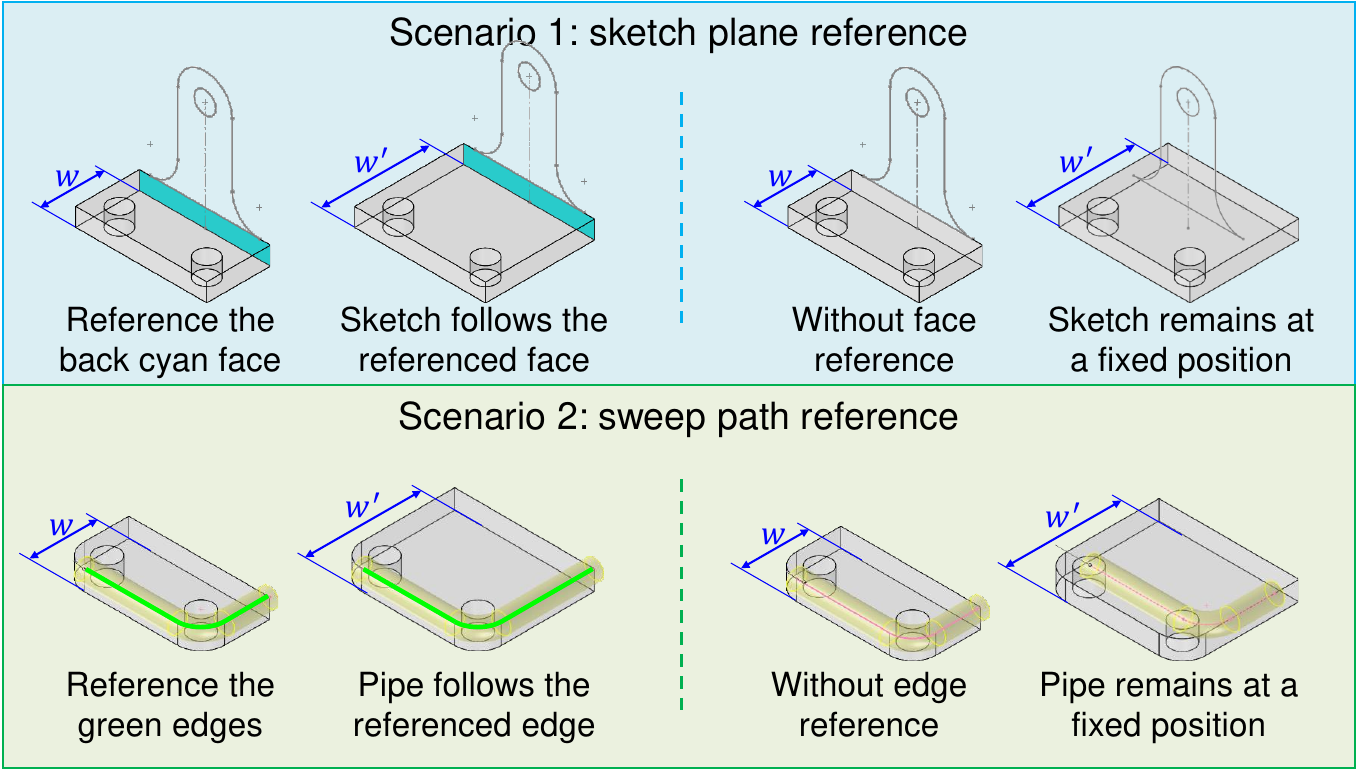}
    \caption{\textbf{Geometry Reference.} With the intended references, after parameter edits \(w\rightarrow w^{'}\), the sketch and swept pipe update with their source geometry. Without those references, the initially identical constructions fail to retain the intended relationships.}
    \label{fig:geom-ref}
\end{figure}

Geometry reference involves selecting specific existing faces/edges/vertices for modeling commands or selecting existing face as sketch plane. For example, creating a swept pipe requires selecting its sweep path. Incorrect references may produce the intended initial shape but lead to unintended updates after parameter edits (\Cref{fig:geom-ref}).

Early methods predict sketch-plane origins and axis directions directly~\citep{wu2021deepcad}, without referencing existing geometry. Consequently, sketch planes cannot automatically adapt to upstream parameter edits. Later approaches add limited dependencies, such as referencing the highest face along +Z direction by descriptor '>Z' \citep{yuan2024openecad,li2026seekcad,cadqueryselector}. Pointer-CAD selects existing B-rep faces or edges by matching predicted pointer embeddings to geometric features, but this similarity-based selection may be ambiguous for geometrically similar entities \citep{qi2026pointercad}. More diverse modeling commands therefore require more general and unambiguous geometric references. Engineers select these entities interactively; an AI system needs a representation that makes them identifiable to the VLM.

\subsection{Sketch Placement (Local Coordinate System)}
\label{subsec:dependency_placement}

\begin{figure}[pos=htbp]
    \centering
    \includegraphics[width=0.65\linewidth]{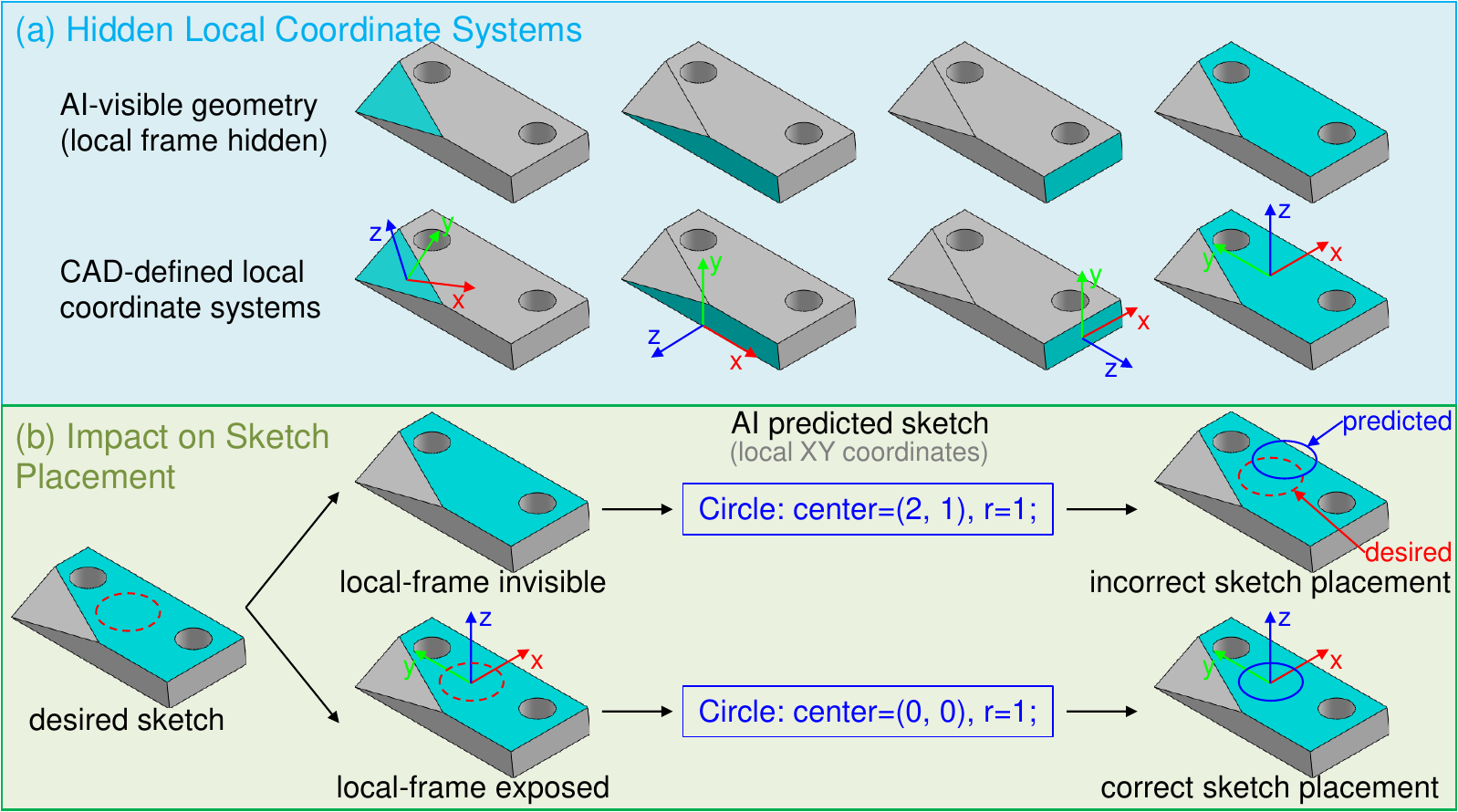}
    \caption{\textbf{Sketch placement.} Although sketch planes can be identified from the visible geometry, their associated local coordinate origins and axis orientations are not explicitly exposed. Without access to these systems, AI may predict incorrect sketch coordinates, resulting in misplaced geometry even when the correct sketch plane has been selected.}
    \label{fig:local-coord-problem}
\end{figure}

The sketch placement problem is how to position the sketch correctly on the sketch plane. Sketch primitives are defined in the sketch plane's local coordinate system. The CAD kernel assigns the origin and axes. Therefore, understanding this local coordinate system is essential for accurately positioning sketch primitives (\Cref{fig:local-coord-problem}).

Existing methods primarily use sketch planes parallel to the XOY, YOZ, or XOZ planes, with local axes aligned with or opposite to the global axes~\citep{wu2021deepcad,wang2024cadgpt,yuan2024openecad}. SkexGen reports that such planes account for over 99\% of DeepCAD sketches~\citep{xu2022skexgen}, reducing the need to handle more varied sketch-plane orientations in that dataset. Nevertheless, inclined sketch planes also occur in practical engineering models. Therefore, these dataset-specific regularities do not eliminate the need to address the general local-coordinate problem. The challenge of sketch placement is how to expose the plane-specific local coordinate systems to the VLM or map predicted coordinates into sketch local coordinates.

\subsection{Sketch External Constraints}
\label{subsec:dependency_constraints}

\begin{figure}[pos=htbp]
    \centering
    \includegraphics[width=0.85\linewidth]{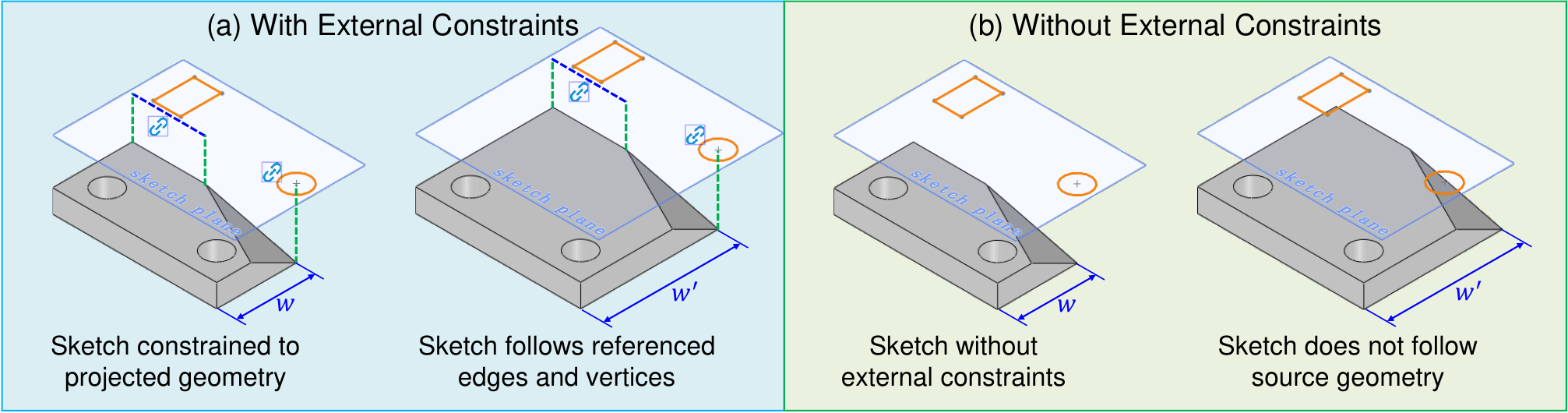}
    \caption{\textbf{Sketch external constraints.} (a) When sketch primitives are constrained to projected edges and vertices, they adapt to modifications of the referenced geometry, preserving the intended design relationships. (b) Without external constraints, the sketch primitives remain at their original local coordinates, potentially violating the intended geometric dependencies after parameter edits.}
    \label{fig:proj-geom-problem}
\end{figure}

Sketch external constraints establish relationships between sketch primitives and existing geometry projected onto the sketch plane. For example, the center of a sketch circle can be constrained to coincide with a projected vertex. This constraint allows the sketch to update automatically when upstream parameters edit (\Cref{fig:proj-geom-problem}).

Several methods output sketch primitive coordinates directly~\citep{wu2021deepcad,xu2022skexgen,yuan2024openecad,li2026seekcad}. Pointer-CAD supports snapping to existing edges~\citep{qi2026pointercad}, but lacks persistent constraints for maintaining such relationships during parameter edits. Establishing Sketch External Constraints requires identifying both the source geometry and its projection on the sketch plane.

\section{Method}
\label{sec:method}

\begin{figure}[pos=htbp]
    \centering
    \includegraphics[width=0.85\linewidth]{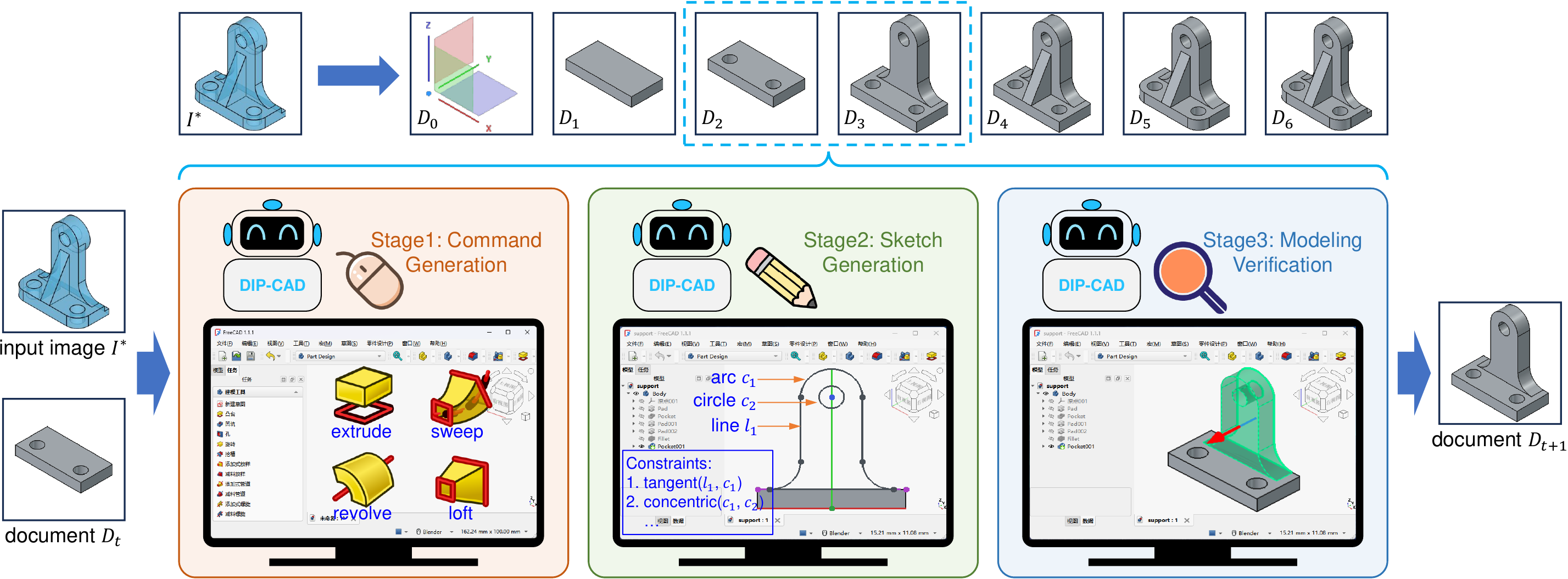}
    \caption{\textbf{\methodname\ overall workflow.} Given a target image \(I^*\), \methodname\ progressively updates an initially empty FreeCAD parametric part document \(D_0\) into the complete document \(D_N\) through a sequence of modeling steps. Each step consists of up to three stages. \textbf{Stage 1: Command Generation} predicts the modeling command, including its command type, geometry references, and parameters. If the predicted command requires a new sketch, the process proceeds to \textbf{Stage 2: Sketch Generation}, which constructs the required parametric sketch; otherwise, Stage 2 is skipped. \textbf{Stage 3: Modeling Verification} generates the modeling preview, and then verified by the VLM. The accepted decision updates \(D_t\) to \(D_{t+1}\); otherwise, the current modeling step is restarted from Stage 1.}
    \label{fig:overall}
\end{figure}

\methodname\ generates a FreeCAD parametric part document from an input image (\Cref{fig:overall}). 
The agent harness tackles the three challenges in \Cref{sec:challenges} by converting CAD data into VLM-friendly formats, as detailed in \Cref{subsec:s1} and \Cref{subsec:s2}.

% with the current CAD state through edge-ID annotations and structured geometry tables, converts predicted coordinates into the selected sketch plane, and automatically constructs and solves supported constraints. 

% The three requirements in \Cref{sec:challenges} guide the DIPA workflow in \Cref{fig:overall}. The harness simplify the three challenges with the current CAD state through edge-ID annotations and structured geometry tables, converts predicted coordinates into the selected sketch plane, and automatically constructs and solves supported constraints. 

% construction that encodes inferred geometric design intent in command references and sketch constraints. Given an input image, the VLM selects operations, reference entities, and approximate profile geometry. 

% Native references and constrained sketches record the resulting dependencies, so that the CAD system can propagate later parameter changes through the feature history.

% Stage~1 generates selects the operation and the geometry on which it depends, and proposes the operation parameters. Stage~2 places the sketch on the selected support and establishes supported geometric and dimensional constraints. Stage~3 executes the candidate operation and uses visual feedback to assess its contribution to the target shape. This verification checks modeling progress; the references and constraints established in the earlier stages govern later parameter propagation.

\subsection{Stage 1: Command Generation}
\label{subsec:s1}

\begin{figure}[pos=htbp]
    \centering
    \includegraphics[width=0.85\linewidth]{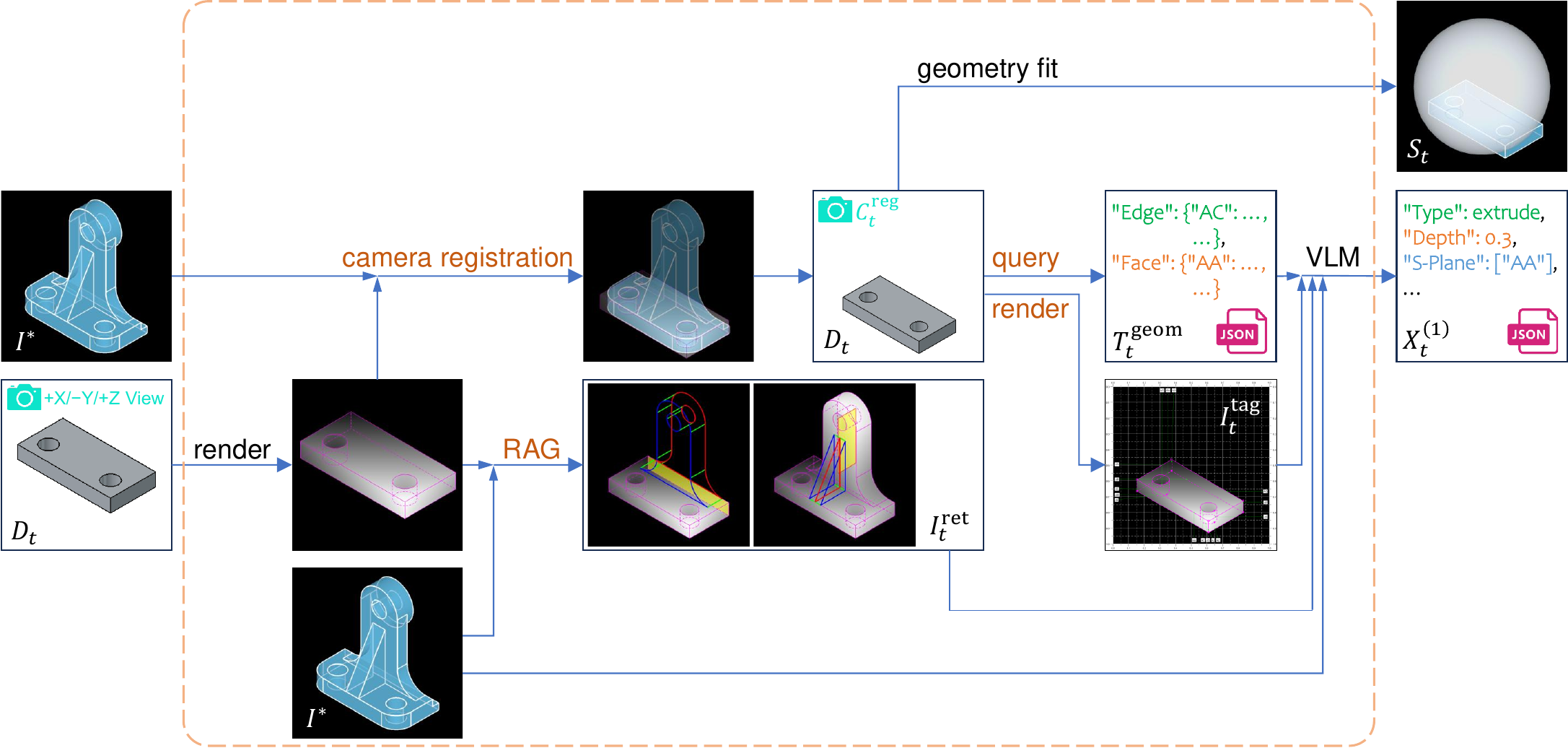}
    % 总标题
    \caption{\textbf{Stage~1: Command Generation workflow.} Camera registration aligns the existing CAD geometry with the target image $I^*$, helping distinguish already constructed geometry from \(I^*\). A pseudo-bounding sphere $S_t$ is also fitted to guide sketch-view framing in Stage 2. Meanwhile, the retrieval-augmented generation (RAG) module retrieves relevant modeling examples to guide the VLM. Under the registered camera $C_t^{\mathrm{reg}}$, a tag image $I_t^{\mathrm{tag}}$ and geometry table $T_t^{\mathrm{geom}}$ are generated (\Cref{fig:tag-img-table}) to enable explicit geometry reference. Given \(I^*\), \(I_t^\mathrm{ret}\), \(I_t^\mathrm{tag}\), and \(T_t^{\mathrm{geom}}\), the VLM predicts the command type, parameters, and referenced geometry IDs.}
    \label{fig:stage1}
\end{figure}

The Stage 1 workflow is illustrated in \Cref{fig:stage1}, where explicit geometry reference is enabled by providing the VLM with the tag image $I_t^{\mathrm{tag}}$ and geometry table $T_t^{\mathrm{geom}}$.

% \subsubsection{Registration}

% We render a depth image $I_t^{\mathrm{depth}}$ from $D_t$. Relative linear depth is shown from white (near) to black (far). Magenta solid and dashed lines mark visible and occluded edges. 
% % The default render view is PNP (view direction: $(-0.577, 0.577, -0.577)$, up vector: $(-0.408, 0.408, 0.816)$).

% Registration adjusts the camera scale and translation to align $I_t^{\mathrm{depth}}$ with $I^*$. The resulting camera $C_t^{\mathrm{reg}}$ places the existing model at the corresponding target location, helping the VLM identify which regions in $I^*$ will be constructed in the current modeling step (\Cref{fig:stage1}).

\subsubsection{Explicit Geometry Reference}
\label{subsubsec:geom_ref}

\begin{figure}[pos=htbp]
    \centering
    \includegraphics[width=0.75\linewidth]{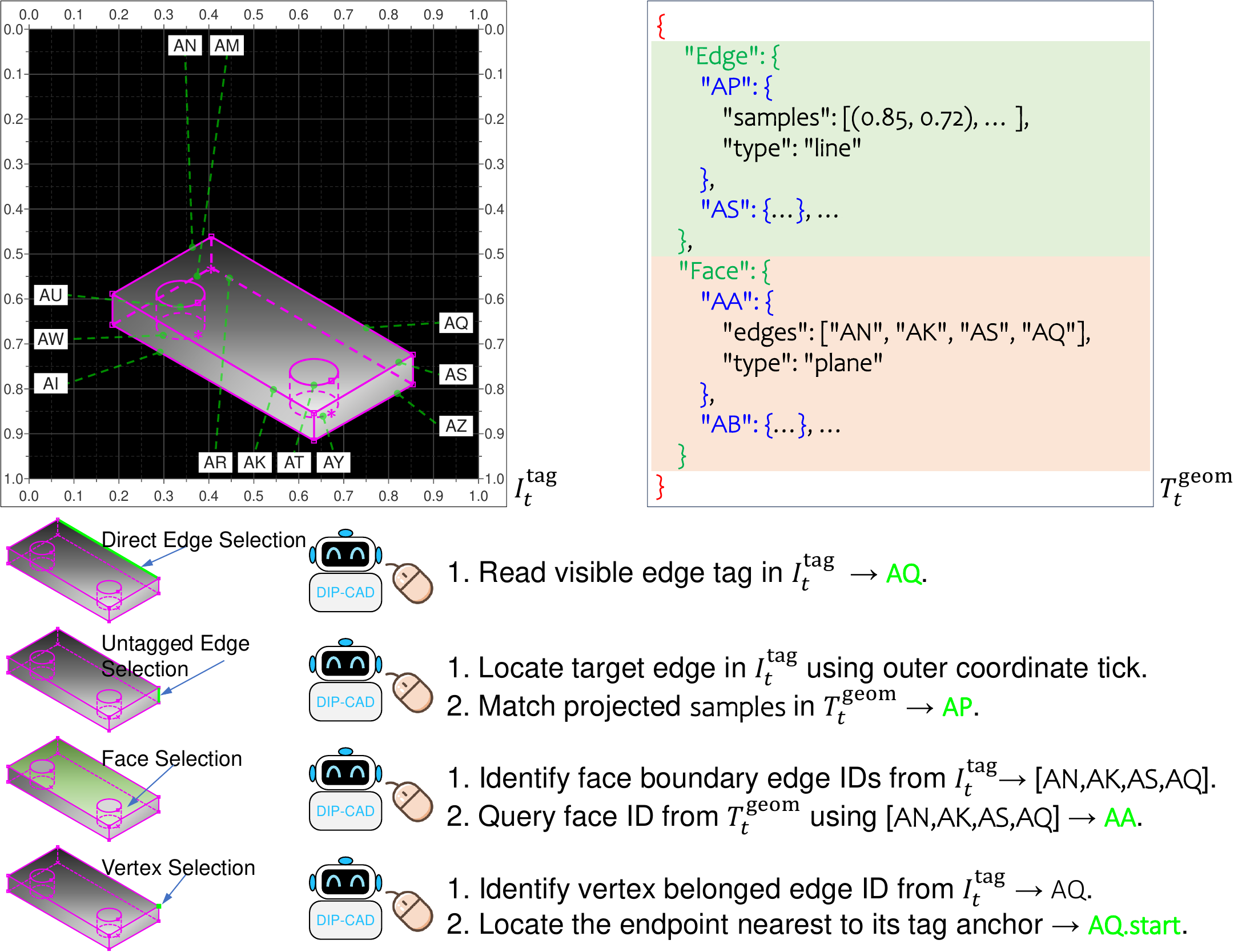}
    % 总标题
    \caption{\textbf{Contents and usage of $I_t^{\mathrm{tag}}$ and $T_t^{\mathrm{geom}}$ in geometry reference.} $I_t^{\mathrm{tag}}$ shows depth, coordinate ticks, and edge ID tags. Visible and occluded edges use solid and dashed magenta lines; vertices use magenta squares and asterisks. ID tags for very short edges and hidden edges after hidden-line removal (HLR) are omitted. $T_t^{\mathrm{geom}}$ lists all edge types and uniformly sampled view coordinates. The VLM selects faces and edges by ID, and vertices by \({ID}_{Edge}.Start\) or \({ID}_{Edge}.End\). Edge samples in $T_t^{\mathrm{geom}}$ are ordered from start to end. Tag leader lines in $I_t^{\mathrm{tag}}$ are anchored at random positions within $[0.1,0.25]$ of the normalized edge parameter domain, typically closer to the start. These cues distinguish the endpoints.}
    \label{fig:tag-img-table}
\end{figure}

We render the tag image $I_t^{\mathrm{tag}}$ and export the geometry table $T_t^{\mathrm{geom}}$ under $C_t^{\mathrm{reg}}$. Partial edge ID tags in $I_t^{\mathrm{tag}}$ and boundary information in $T_t^{\mathrm{geom}}$ help the VLM identify target geometry explicitly (\Cref{fig:tag-img-table}). 

\(I_t^{\mathrm{tag}}\) and \(T_t^{\mathrm{geom}}\) provide complementary information for geometry reference. \(I_t^{\mathrm{tag}}\) conveys the geometry spatial locations through rendered edges and edge-ID tags, while \(T_t^{\mathrm{geom}}\) lists all geometry IDs and provides view-space sampled points for edge localization. Their shared IDs and common view coordinate system jointly enable complete reference of faces, edges, and vertices.

\subsubsection{Retrieval-Augmented Generation (RAG)}

We use the current depth image \(I_t^{\mathrm{depth}}\) together with the target image \(I^*\) to retrieve similar modeling examples. Each retrieved example contains a pseudo-preview \(I_t^{\mathrm{ret}}\) and its corresponding command type \(o_t^{\mathrm{ret}}\). These examples provide additional modeling knowledge that helps the VLM infer the appropriate modeling command.

Similarity between \(\left(I^{\mathrm{depth}},I^*\right)\) and \(I_t^{\mathrm{ret}}\) is a weighted combination of cosine similarities:
\[
\begin{aligned}
s_{i}^{\mathrm{depth}}
&=
\phi\!\left(I_t^{\mathrm{depth}}\right)^{\mathsf T}
\phi\!\left(I_i^{\mathrm{depth}}\right),
\\[2pt]
s_{i}^{\mathrm{target}}
&=
\phi\!\left(I^{*}\right)^{\mathsf T}
\phi\!\left(I_i^{\mathrm{*}}\right),
\\[2pt]
s_i
&=
\lambda s_{i}^{\mathrm{depth}}
+
(1-\lambda)s_{i}^{\mathrm{target}},
\qquad
0\leq\lambda\leq1.
\end{aligned}
\]

Here, $s_i^{\mathrm{depth}}$ and $s_i^{\mathrm{target}}$ measure similarities between current model images and between target model images, respectively. The weight $\lambda$ balances the two. $\phi(I)$ is the normalized [CLS] token from a frozen DINOv3 ViT-B/16 backbone~\citep{simeoni2025dinov3}.

\subsubsection{VLM Invocation}

The Stage~1 VLM input--output mapping is
\[
X_t^{(1)}
=
\operatorname{VLM}\left(
I^{*},
I_t^{\mathrm{tag}},
T_t^{\mathrm{geom}},
I_t^{\mathrm{ret}},
o_t^{\mathrm{ret}},
\mathcal{P}^{(1)}
\right)
\]

The prompt $\mathcal{P}^{(1)}$ explains the inputs and requests the command type, references, and parameters. The output $X_t^{(1)}$ is structured as a JSON object containing seven main fields.
% : TYPE, REQUIRE\_SKETCH, SKETCH\_REF, SKETCH\_HINT, OTHER\_REFERENCE, PARAMETER\_HINT, DIR\_HINT.

The TYPE field specifies the modeling command type, selected from 16 predefined options, including additive and subtractive extrusions, revolutions, sweeps, and lofts, as well as symmetric extrusions and revolutions, fillet, chamfer, construction-plane, and END for terminating the modeling process.

The REQUIRE\_SKETCH field indicates whether a new sketch is required. If true, SKETCH\_PLN identifies the sketch plane ID, while SKETCH\_HINT describes the intended sketch geometry to guide Stage~2. Otherwise, Stage~2 is skipped. 

The OTHER\_REFERENCE field specifies additional geometric references, such as revolution edge ID, fillet or chamfer edge ID. 
PARAMETER\_HINT provides approximate command parameters, such as extrusion depths and revolution angles. Parameters are estimated relative to the $I_t^{\mathrm{tag}}$ view scale.
The DIR\_HINT field describes the intended construction direction in $I_t^{\mathrm{tag}}$ view.

% \begin{table*}[pos=t]
% \centering
% \caption{\textbf{Main fields of the Stage~1 output $X_t^{(1)}$.} The \texttt{TYPE} field accepts 16 values: extrude\_add, extrude\_cut, symmetric\_extrude\_add, symmetric\_extrude\_cut, revolve\_add, revolve\_cut, symmetric\_revolve\_add, symmetric\_revolve\_cut, sweep\_add, sweep\_cut, loft\_add, loft\_cut, fillet, chamfer, cplane\_offset, and END.}
% \label{tab:stage1-json-schema}
% \begin{tabular}{p{0.26\textwidth}p{0.66\textwidth}}
% \hline
% Field & Description \\
% \hline
% \texttt{TYPE} & Type of the next modeling command, or the termination token \texttt{END}. \\
% \texttt{REQUIRE\_SKETCH} & Whether next modeling command requires reference a new sketch. Stage~2 is skipped if this value is \texttt{false}. \\
% \texttt{SKETCH\_REF} & Sketch plane ID for the new sketch, provide sketch plane for Stage 2 sketching. \\
% \texttt{SKETCH\_HINT} & Brief description of the intended shape and location, guiding Stage~2 sketching. \\
% \texttt{OTHER\_REFERENCE} & Other reference IDs: directions, revolution axes, sweep paths, loft sections, fillet/chamfer edges, and construction-plane source faces. \\
% \texttt{PARAMETER\_HINT} & Approximate depth, angle, offset, fillet radius, or chamfer distance, relative to the view scale in $I_t^{\mathrm{tag}}$. \\
% \texttt{DIR\_HINT} & Qualitative direction in the current view, such as extrusion along the positive or negative side of the sketch plane. \\
% \hline
% \end{tabular}
% \end{table*}

\subsection{Stage 2: Sketch Generation}
\label{subsec:s2}

\begin{figure}[pos=htbp]
    \centering
    \includegraphics[width=0.75\linewidth]{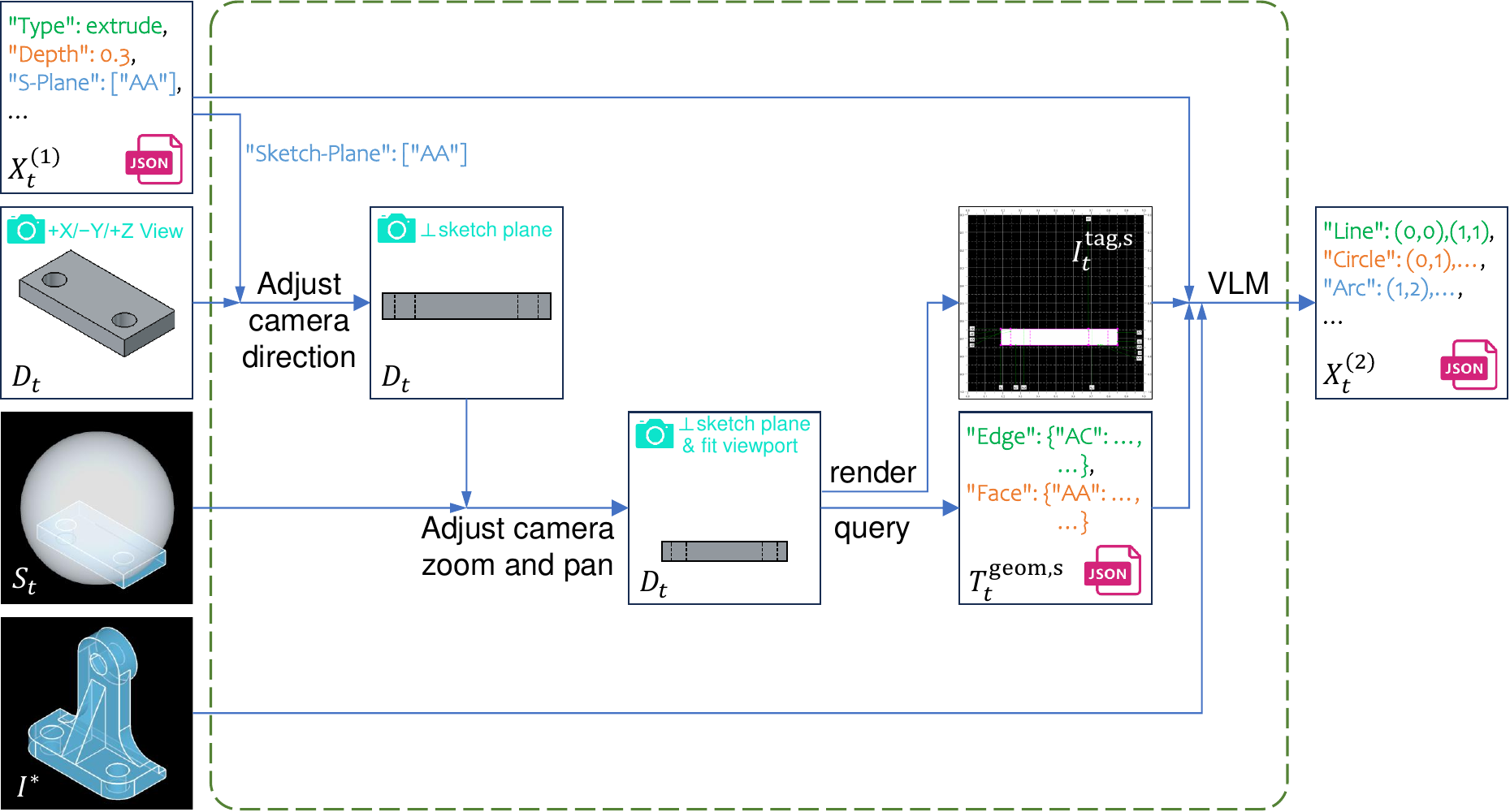}
    % 总标题
    \caption{\textbf{Stage~2: Sketch Generation workflow.} 
    The normal of sketch plane specified in $X_t^{(1)}$ determines the camera orientation. The pseudo-sphere $S_t$ is then used to set the camera's zoom and panning parameters, thereby establishing a suitable sketch viewport (details in Supplementary). From this view, we render the tag image $I_t^{\mathrm{tag,s}}$ and export the associated geometry table $T_t^{\mathrm{geom,s}}$. $I_t^{\mathrm{tag,s}}$ provides sketch view coordinate system, while $T_t^{\mathrm{geom,s}}$ supplies projected external geometry coordinates to ground the VLM's predictions. The VLM takes these inputs to predict the sketch-primitive representation $X_t^{(2)}$. The predicted view coordinates are subsequently transformed into the sketch plane's local coordinate system, and the corresponding geometric constraints are automatically applied.}
    \label{fig:stage2}
\end{figure}

Stage 2 is used to generate the parametric sketch required for the Stage 1 predicted modeling command. The sketch must be correctly placed on the sketch plane and include dimensional constraints, internal constraints and external constraints (\Cref{fig:paramcad-ele}). The Stage 2 workflow is shown in \Cref{fig:stage2}.

\subsubsection{Data Preparation for Sketch Placement}

% \subsubsection{Coordinate bridge for sketch placement}

Accurate sketch placement requires the VLM to identify both the local coordinate system of the sketch plane and the positions of existing geometry projected on the sketch plane. To address this issue, we prepare two complementary inputs for Stage~2: a sketch-oriented view tag image $I_t^{\mathrm{tag,s}}$ and a corresponding geometry table $T_t^{\mathrm{geom,s}}$ (\Cref{fig:stage2}).

% To obtain $I_t^{\mathrm{tag,s}}$ and $T_t^{\mathrm{geom,s}}$, we first orient the camera normal to the sketch plane. The pseudo-bounding sphere \(S_t\) is then used to determine an appropriate camera translation and zoom, ensuring that the sketching area remains largely unobstructed. Subsequently, tag image $I_t^{\mathrm{tag,s}}$ and geometry table $T_t^{\mathrm{geom,s}}$ are rendered using the same representation as in Stage~1, but under the updated camera. 

\begin{figure}[pos=htbp]
    \centering
    \includegraphics[width=0.75\linewidth]{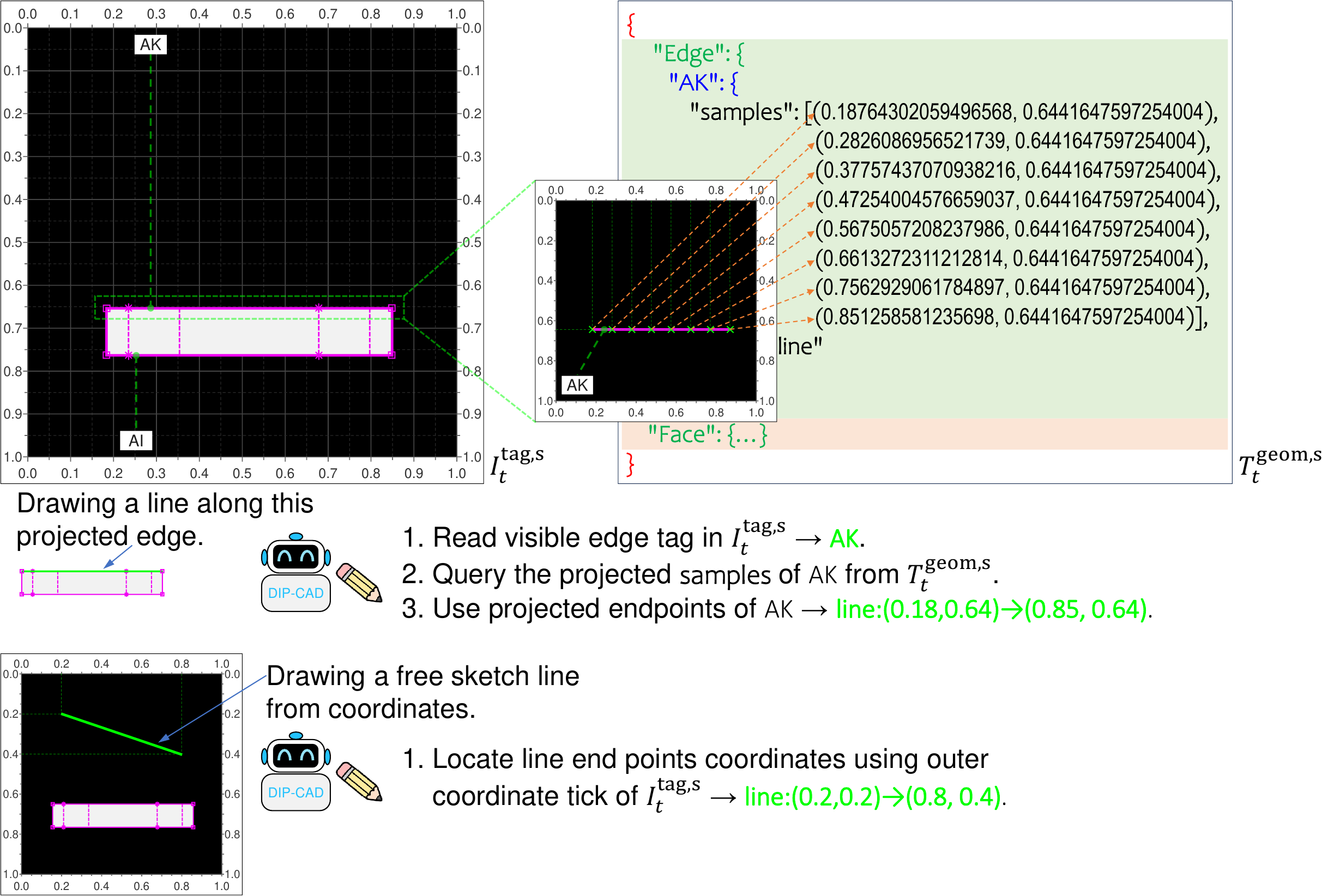}
    % 总标题
    \caption{\textbf{Contents and usage of \(I_t^{\mathrm{tag,s}}\) and \(T_t^{\mathrm{geom,s}}\) in sketching.} The sketch-oriented view tagged image \(I_t^{\mathrm{tag,s}}\) provides a normalized 2D view with edge tags and outer coordinate ticks, enabling direct coordinate estimation. The geometry table \(T_t^{\mathrm{geom,s}}\) stores structured projected geometry, such as projected edge sampled 2D points. Together, they support two typical sketching modes: (1) referencing an existing projected edge to construct sketch primitives, and (2) directly creating new sketch primitives from normalized coordinates.}
    \label{fig:tag-img-s}
\end{figure}

The coordinate ticks around $I_t^{\mathrm{tag,s}}$ explicitly define a normalized view coordinate system, in which the VLM predicts the coordinates of sketch primitive coordinates. Meanwhile, $T_t^{\mathrm{geom,s}}$ provides sampled points on every projected external edge, expressed in the same normalized view coordinates. These samples help locate projected external geometry. For example, if a sketch line should coincide with a projected external edge, the VLM can directly use the projected coordinates of the sampled edge endpoints as the start and end points of the sketch line (\Cref{fig:tag-img-s}).

\subsubsection{VLM Invocation and Coordinate Bridge}

The Stage~2 VLM input--output mapping is
\[
X_t^{(2)}
=\operatorname{VLM}
\left[
I^{*},
I_t^{\mathrm{tag,s}},
T_t^{\mathrm{geom,s}},
X_t^{(1)},
\mathcal{P}^{(2)}
\right]
\]

The prompt $\mathcal{P}^{(2)}$ requests one or more closed loops in the view coordinates of $I_t^{\mathrm{tag,s}}$. Edge IDs connect the visible reference geometry to projected samples in $T_t^{\mathrm{geom,s}}$. The output $X_t^{(2)}$ defines primitives in JSON: a line uses two endpoints; an arc uses start, intermediate, and end points; and a circle uses four ordered points on its circumference.

After $X_t^{(2)}$ is obtained, the CAD kernel maps the VLM-predicted normalized view coordinates to the sketch plane local coordinate system. This design avoids requiring the VLM to infer the CAD kernel-specific definition of the sketch-local coordinate system, while still producing precise coordinates for native sketch construction.

% The harness converts these predicted view coordinates into the selected sketch plane's native local coordinates using the camera and sketch-plane coordinate systems. This deterministic transformation handles the difference in origin, orientation, and scale between the rendered view and the sketch. Entity labels and projected samples guide where the profile belongs, while coordinate conversion instantiates that placement in the CAD document. The resulting primitives provide the geometry on which automatic constraint generation operates.

\subsubsection{Sketch Constraint Generation}

% This step implements projected geometry referencing by automatically imposing native sketch constraints. The projected edge samples used for placement describe the current geometry; a persistent link additionally requires a reference to the source entity and a constraint relating the sketch primitive to its projection.
% \methodname\ detects constraints using tolerance tests. Constraints are applied in the following order: External Constraints Generation, Internal Constraints Generation, and Dimensional constraints Generation.
% This process converts plausible geometric intent into rules that the sketch solver can maintain during editing.

To preserve sketch primitive relationships during parameter editing, \methodname\ automatically generates sketch constraints. Candidate constraints are identified through tolerance-based geometric tests, including positional, angular, length, and radius comparisons. Constraints are applied following a predefined priority order.

(1) External Constraint Generation.
% Existing edges are imported as external sketch geometry using their ID and projected coordinates. For sketch lines and arcs, we apply constraints in the following order: (i) collinearity between sketch line and projected line; (ii) coincidence of sketch line or arc endpoint with a projected edge endpoint or the projected circle center; and (iii) point-on-object constraints that place a sketch line or arc endpoint on a projected edge. For circular profiles, the system also automatically establishes center coincidence and equal-radius constraints between a sketch circle and a projected circular edge when the corresponding tolerance tests are satisfied. These constraints preserve the sketch circle's center and radius relationships to its source during later edits, as illustrated in \Cref{subsec:mechanism_case}.
Existing model edges are first projected onto the sketch plane, and their source IDs are also recorded. Constraints between sketch primitives and projected external geometry are then applied in the following order: (i) collinearity between lines; (ii) coincidence between key points; and (iii) key points on object; and (iv) equal radii/lengths. Key point includes primitive endpoint and circle/arc center.

(2) Internal Constraint Generation.
% We next apply endpoint coincidence, horizontal and vertical line constraints, tangency involving circles or arcs, equal line lengths, equal arc radii, and point symmetry, in that order. Each candidate is tested using the positional, angular, length, or radius adjustment needed to satisfy the relationship.
% For symmetry constraints, we project the outer boundary of the sketch-support face. If the projected outline is symmetric about both its horizontal and vertical midlines, their intersection and the two midlines define the reference center and symmetry axes. Otherwise, we use the sketch origin and its $X/Y$ axes. The same fallback applies when no supporting face is available. Symmetry constraints are detected on primitive endpoints and circle or arc centers.
Constraints among sketch primitives are applied in the following order: (i) coincidence between key points; (ii) horizontal and vertical alignment; (iii) tangency involving lines, circles, and arcs; (iv) equal radii/lengths; and (v) key points symmetry. Symmetry references are derived from the projected support-face boundary when it is symmetric about both midlines; otherwise, the sketch origin and axes are used.

(3) Dimensional Constraint Generation.
Finally, dimensional constraints are added to specify line lengths, circle or arc radii, and key point offsets relative to the reference center and axes.

Each candidate constraint is validated by the FreeCAD sketch solver. Constraints that would over constrain the sketch are rejected, and the process terminates when the sketch is fully constrained or no further valid constraints can be added.

\subsection{Stage 3: Modeling Verification}
\label{subsec:s3}

Stage~3 checks whether the predicted result would bring the existing model closer to the target, as shown in \Cref{fig:stage3}. 
% FreeCAD visualizes the command defined by $X_t^{(1)}$ and, when required, $X_t^{(2)}$ by rendering a modeling preview $I_t^{\mathrm{pv}}$. 
The Stage~3 VLM input-output mapping is: 

% The candidate is executed for preview before it is accepted into the feature history. A rejected candidate leaves $D_t$ unchanged and returns the modeling step to Stage~1 with feedback.

\begin{figure}[pos=htbp]
    \centering
    \includegraphics[width=0.6\linewidth]{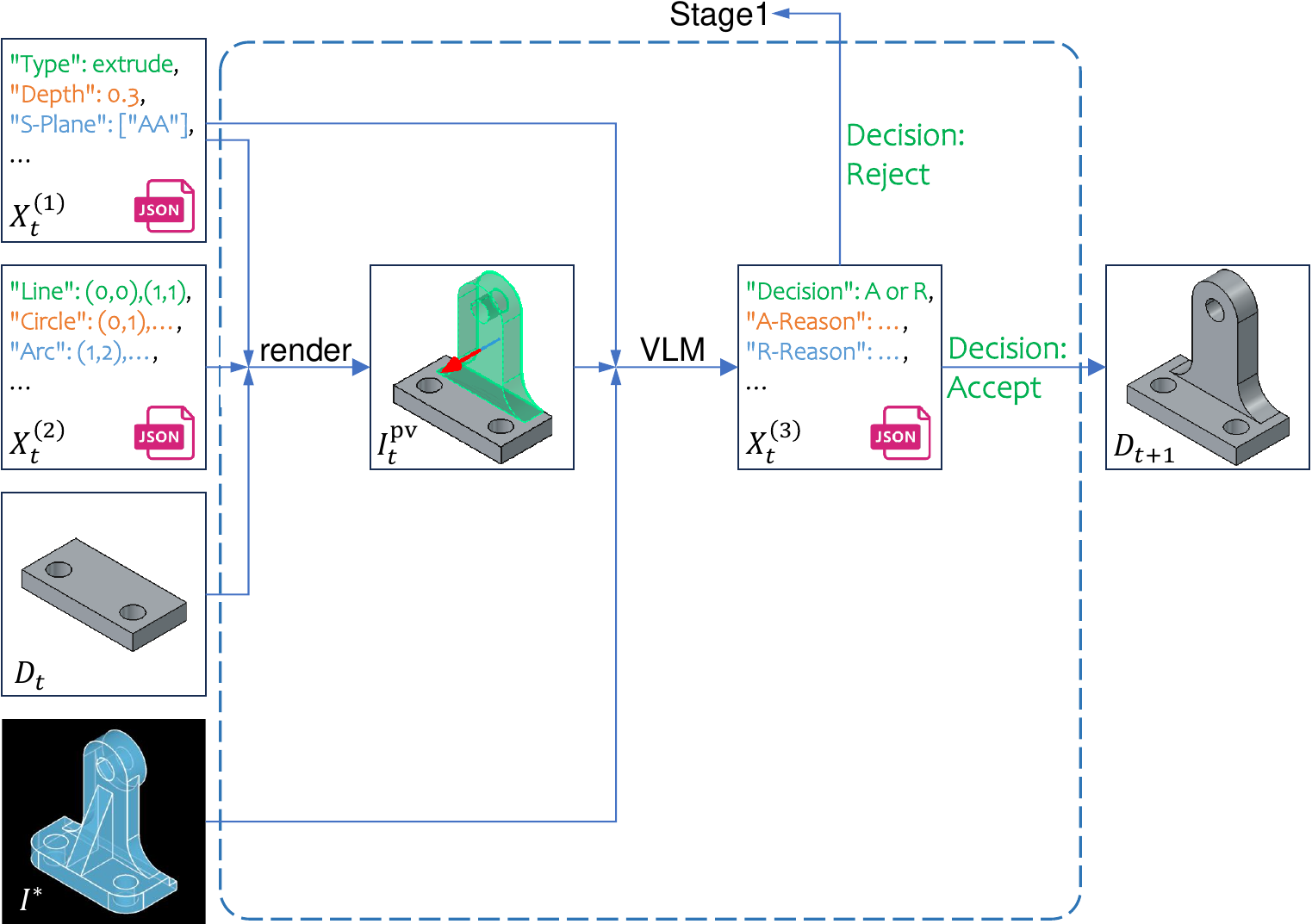}
    % 总标题
    \caption{\textbf{Stage~3: Modeling Verification workflow.} FreeCAD executes $X_t^{(1)}$ and, when needed, the sketch derived from $X_t^{(2)}$ against $D_t$ to render $I_t^{\mathrm{pv}}$. The VLM evaluates the candidate using $X_t^{(1)}$, $I_t^{\mathrm{pv}}$, and the target $I^*$, producing a decision $X_t^{(3)}$. Acceptance commits the command and produces $D_{t+1}$; rejection returns to Stage~1 with reasons. The VLM context is retained across retries within the same modeling step and reset after an accepted step.}
    \label{fig:stage3}
\end{figure}

\[
X_t^{(3)}
=\operatorname{VLM}
\left[
X_t^{(1)},
I^{*},
I_t^{\mathrm{pv}},
\mathcal{P}^{(3)}
\right]
\]

The prompt $\mathcal{P}^{(3)}$ asks whether the candidate creates a missing target structure or a useful intermediate feature. The JSON output $X_t^{(3)}$ contains a decision and a brief acceptance justification or detailed rejection reasons.

\section{Experiments}
\label{sec:experiments}

We evaluate design-intent-preserving CAD generation in terms of both static shape accuracy and parameter-editing performance. 
% We examine whether \methodname\ matches the target geometry, whether command references and sketch constraints preserve the encoded relationships during editing, and which mechanisms contribute to these outcomes. 
Static metrics quantify counter fidelity, editing scores assess the response to a predefined parameter edit, and a representative case traces the underlying dependencies. 
% Ablations examine the roles of geometric information, coordinate conversion, and constraint construction within the generation workflow.

\subsection{Geometry Explicit Reference Dataset (GERD)}
\label{subsec:gerd}

Existing CAD datasets often lack step-level annotations that explicitly link modeling commands to their referenced geometry. Moreover, native parametric CAD file formats are incompatible across CAD systems. These limitations hinder the study of geometry reference across modeling steps. To address this gap, we design a CAD-system-independent data representation with explicit geometry reference (details in supplementary).
Based on this representation, we construct the Geometry Explicit Reference Dataset (GERD) from 2,700 publicly available Onshape parts. The dataset contains 2,500 training, 100 validation, and 100 test samples. Supported command types include extrude, revolve, sweep, loft, chamfer, fillet, and construction-plane.

% GERD-RET uses the GERD training split for modeling example retrieval (\Cref{fig:ret-data}). GERD-EVL is derived from GERD for image-to-CAD code generation and inherits its training, validation, and test splits. Each sample in GERD-EVL is $\{I_i,C_i^{\mathrm{fs}},M_i^{\mathrm{step}},M_i^{\mathrm{ex10p}}\}$. Here, $I_i$ is the completed-part image and $C_i^{\mathrm{fs}}$ is its cadfs modeling program~\citep{pyatov2026cadfs}. The two STEP files \(M_i^{\mathrm{step}},M_i^{\mathrm{ex10p}}\) store the original geometry and the result after increasing all extrusion depths by $10\%$. 

% GERD also supports the derivation of datasets with different viewpoints, rendering modalities, and supervision targets. Potential tasks include next-command prediction, face/edge/vertex selection, B-Rep encoding, and validating references across modeling steps. GERD thus serves both as a data source for this work and as a reusable intermediate representation for stepwise parametric CAD modeling.

GERD supports the construction of datasets for different parametric CAD modeling tasks. In this work, we derive two task-specific datasets. GERD-RET uses the GERD training split as a retrieval database of modeling examples. GERD-EVL supports image-to-CAD code generation, following the original training, validation, and test splits. Each instance in GERD-EVL is represented as $\{I_i^*,C_i^{\mathrm{fs}},M_i^{\mathrm{step}},M_i^{\mathrm{ex10p}}\}$. Where $I_i^*$ is the completed-part image, and $C_i^{\mathrm{fs}}$ is the corresponding CADFS code~\citep{pyatov2026cadfs}. The two STEP files, $M_i^{\mathrm{step}}$ and $M_i^{\mathrm{ex10p}}$, contain the original model and the edited model obtained by increasing all extrude depths by $10\%$, respectively. 
% The latter enables evaluation of model updates following parameter edits.

\subsection{Evaluation Metrics}
\label{subsec:evaluation_metrics}

% The static shape and its response to an edit provide complementary evidence. A model can match the target geometry while relying on independent coordinates that fail to preserve an intended relationship when a source feature changes. We therefore evaluate each generated model both before and after a controlled parameter change.

We report compilation success rate (CR), mean intersection over union (mIoU), Chamfer distance (CD), Hausdorff distance (HD), and 95th-percentile Hausdorff distance (HD95). For \methodname, CR counts generated FreeCAD documents that are non-empty and report no errors. Geometric metrics are computed only for outputs counted as successful under the CR criterion, and must be interpreted alongside CR.

For the editing protocol, all additive and subtractive extrusion depths are simultaneously multiplied by $1.1$. We then recompute CR and the geometric metrics against the edited ground truth. This tests whether the model remains executable and how closely its updated geometry matches the edited ground truth after the same relative depth change.

% Multiple valid modeling histories can produce the same part, so matching every generated reference or constraint to one ground-truth history is not generally meaningful. The editing protocol instead provides an indirect assessment of the encoded dependencies. Its scores also reflect initial reconstruction error and differences in construction history: the same relative depth change need not have the same semantic effect in different histories. We therefore interpret these scores together with the dependency traces in the case study and the controlled ablations, rather than as a complete measure of design-intent preservation under arbitrary edits.

\subsection{Static Shape Accuracy and Parameter Editing Performance}
\label{subsec:main_results}

\begin{figure}[pos=htbp]
    \centering
    \includegraphics[width=1.0\linewidth]{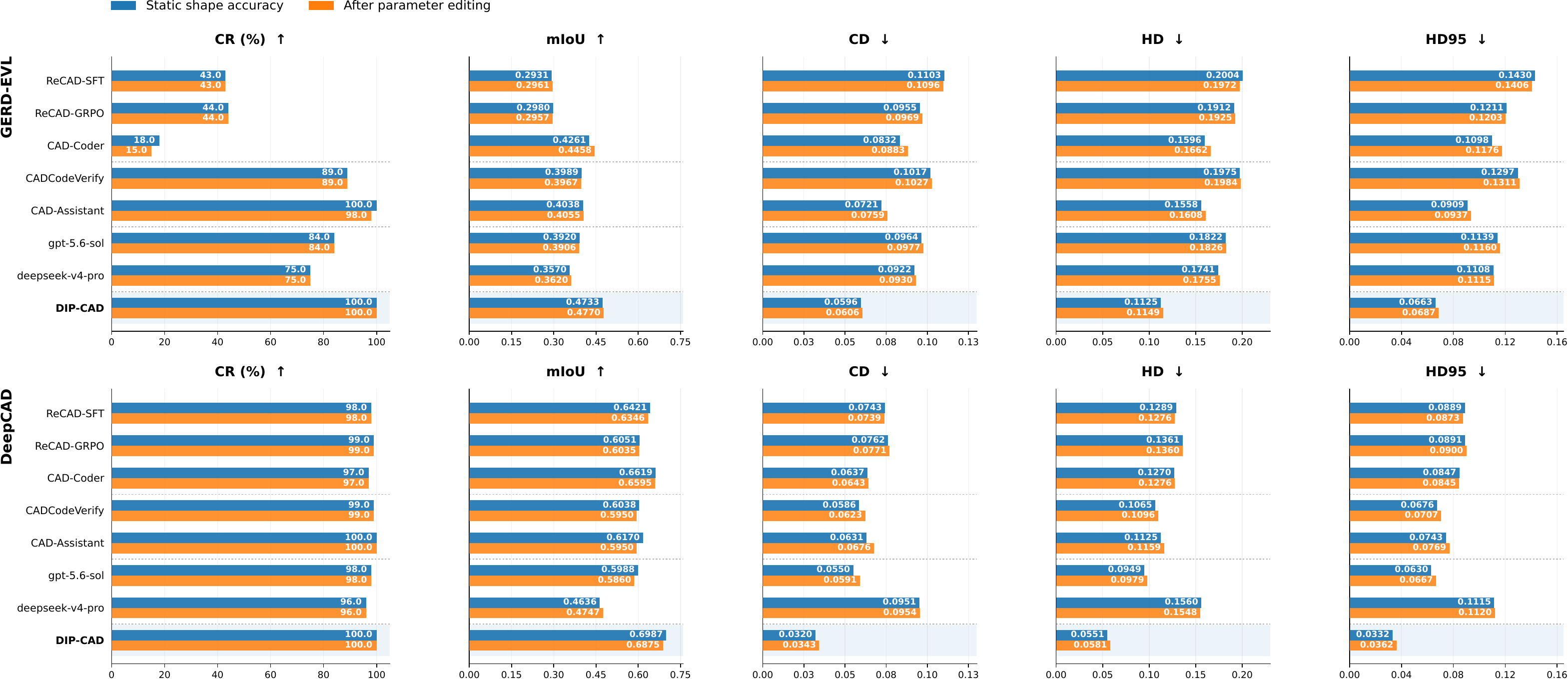}
    \caption{\textbf{Static shape accuracies and parameter edit performances.}}
    \label{fig:exp_res}
\end{figure}

Static shape accuracies are shown in \Cref{fig:exp_res} blue bar, in which \methodname\ returns non-empty documents with no reported errors for all test cases ($100\%$ CR) and achieves the best geometric metrics on both datasets. On GERD-EVL, mIoU improves by $11.1\%$ and CD decreases by $17.3\%$ relative to the strongest baseline for each metric. These comparisons are based on each method's outputs that satisfy the CR criterion: CAD-Coder's mIoU of $0.4261$ covers only 18 of the 100 test parts, whereas \methodname's scores cover all 100.
DeepCAD dataset mainly contains regular sketch-extrude sequences. Most baselines reach at least $97\%$ CR, so the evaluated subsets differ less in size than on GERD-EVL. \methodname\ improves mIoU by $5.6\%$ and reduces CD by $41.8\%$ relative to the strongest baseline for each metric. These results indicate \methodname\ improved static shape accuracy on the both evaluated datasets.

Parameter editing performances are shown in \Cref{fig:exp_res} orange bar. \methodname\ retains $100\%$ CR and the best reported geometric metrics after the uniform extrude-depth increase. Relative to the strongest baseline for each metric, mIoU improves by $7.0\%$ on GERD-EVL and $4.2\%$ on DeepCAD. CD decreases by $20.2\%$ and $42.0\%$, respectively. These scores show closer agreement with the edited ground truth across \methodname's evaluated outputs.

\begin{figure}[pos=t]
    \centering
    \includegraphics[width=0.75\linewidth]{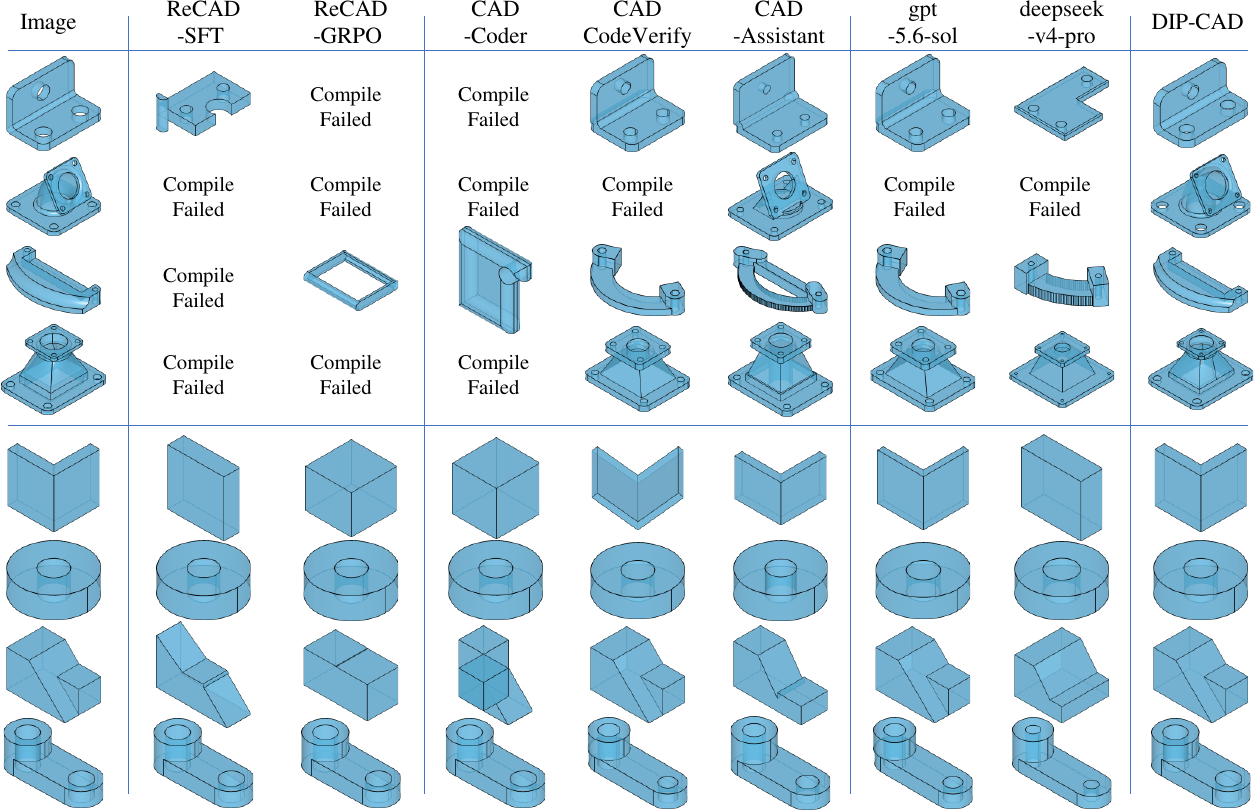}
    \caption{\textbf{Generated model examples.} The leftmost column shows the target image, the remaining columns show the generated models. The upper four rows are GERD-EVL examples; the lower four are from a DeepCAD test subset. \methodname\ more consistently recovers the required supports, holes, and curved profiles.}
    \label{fig:qualitative_comparison}
\end{figure}

The generated examples are shown in \Cref{fig:qualitative_comparison}.
In the upper four GERD-EVL examples, baseline outputs miss holes, supports, or curved profiles. Most methods recover the basic shapes of the DeepCAD examples, while \methodname\ better matches their local contours and feature positions.

 % Entries labeled ``Compile Failed'' indicate unsuccessful model generation or execution. \methodname\ more consistently recovers the required supports, holes, and curved profiles.

\subsection{Case Study on Design-Intent Preservation}
\label{subsec:mechanism_case}

Although static reconstruction and predefined parameter-editing evaluations measure geometric accuracy before and after parameter edits, they cannot fully reveal whether the generated models preserve the underlying design intent. Moreover, the construction of a CAD model is not unique, as the same target geometry can be obtained through different modeling sequences and geometric references. Consequently, evaluating stepwise command prediction and geometry reference accuracy against a single ground-truth may not reliably reflect modeling correctness. To address these limitations, we conduct a qualitative case study to access the design-intent-preserving performance. The case study examines the generated feature history, geometric dependencies, and their behavior under parameter edits.

\begin{figure}[pos=t]
    \centering
    \includegraphics[width=1.0\linewidth]{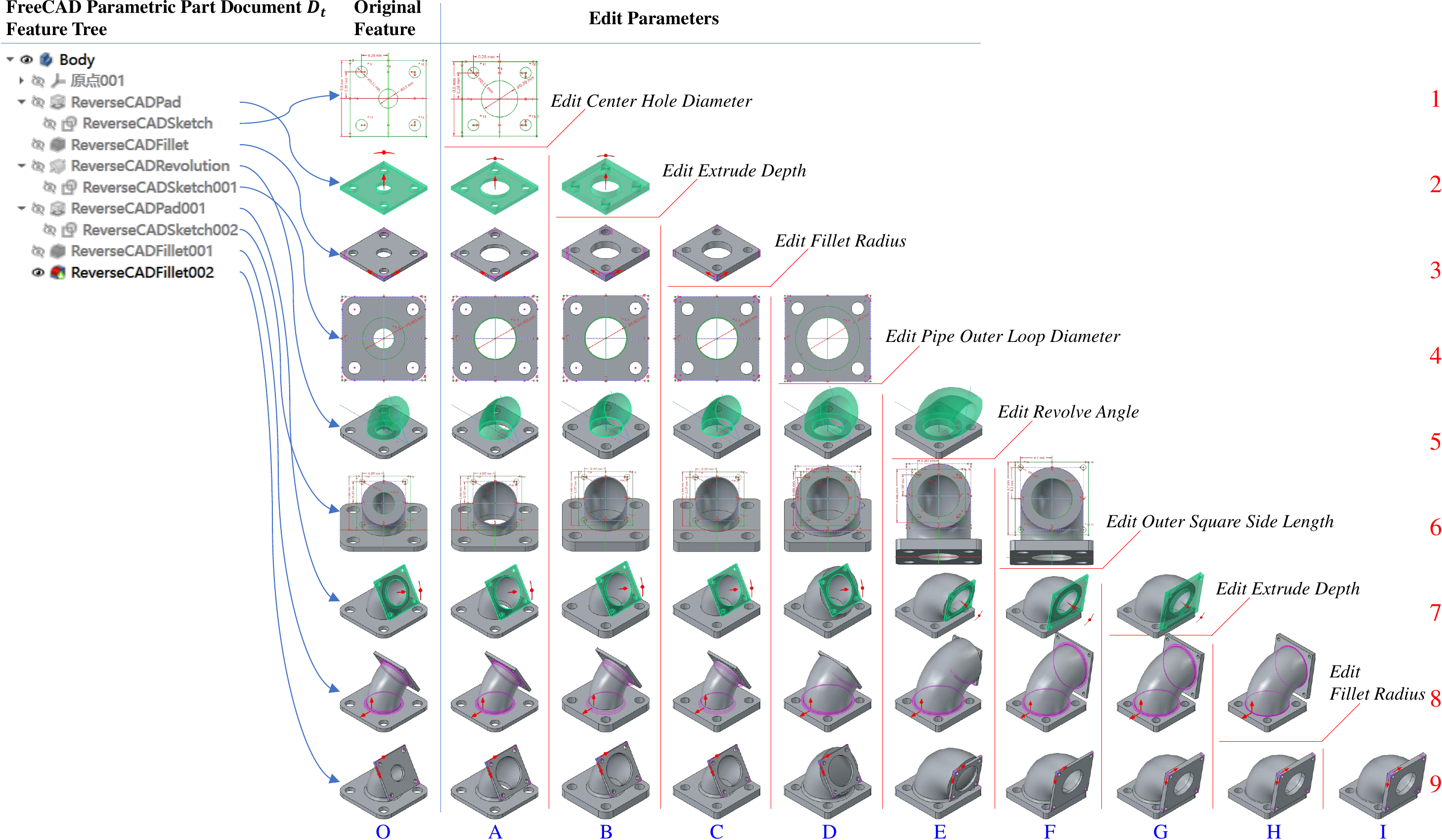}
    \caption{\textbf{Parameter edit propagation case.} Blue arrows link the feature tree to the corresponding modeling preview. Column O shows the original states; columns A–I show previews obtained during parameter editing. Rows 1-9 denote stages in the feature tree. Column-row identifiers (e.g., O3 and A3) are also used in \Cref{fig:mechanism_cases}.}
    \label{fig:parameter_propagation}
\end{figure}

\Cref{fig:parameter_propagation} illustrates the feature tree of a representative DIP-CAD-generated FreeCAD model and the propagation of successive parameter edits. Starting from the original model (column O), we modify various parameters, including sketch dimensions, extrude depths, fillet radii, and revolve angles (columns A-I). As the parameters edit, the affected features and their downstream geometry update accordingly, without visible disconnections, unintended feature detachment, or structural failures. These results demonstrate that the generated model supports coherent parameter-driven updates across multiple modeling commands.

\begin{figure}[pos=htbp]
    \centering
    \includegraphics[width=1.0\linewidth]{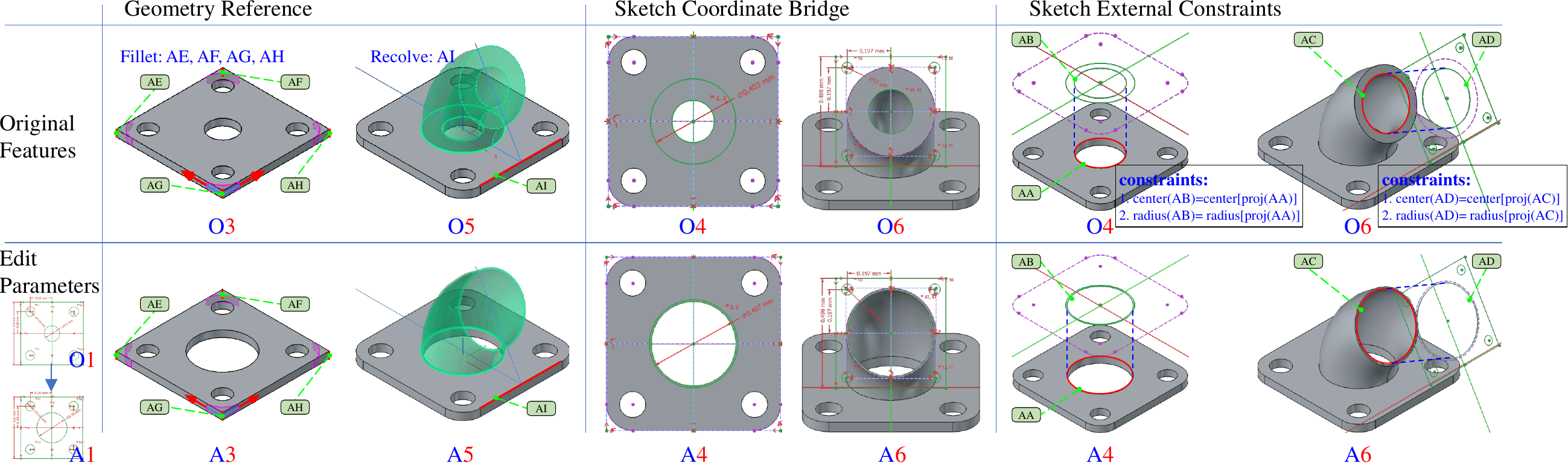}
    \caption{\textbf{Mechanism-level analysis of parameter-edit propagation.} 
    The upper and lower rows show the original and updated features following an increase in the center-hole diameter (O1 $\rightarrow$ A1), respectively.
    \textbf{Left:} Explicit geometric references for fillet and revolve commands.
    \textbf{Middle:} Sketch placement on different support planes through the coordinate bridge.
    \textbf{Right:} Persistent constraints between sketch primitives and projected external geometry (Sketches in the right panels are offset from their sketch planes for visibility only).}
    \label{fig:mechanism_cases}
\end{figure}

To further investigate how design intent is preserved, \Cref{fig:mechanism_cases} examines representative downstream features from \Cref{fig:parameter_propagation} following an increase in the center-hole diameter (O1 $\rightarrow$ A1). First, for geometry reference, the fillet in O3 references edges AE, AF, AG, and AH, while the revolution in O5 references edge AI as its axis. These references remain valid after the upstream edit (A3 and A5), demonstrating stable geometric entity referencing. Second, for sketch placement, O4 and O6 show sketches constructed on different support planes. The sketch coordinate bridge maps VLM-predicted view coordinates to the native local coordinate systems, enabling correct placement of sketch primitives on their respective planes. Third, for sketch external constraints, sketch circles are constrained to the projections of source edges, respectively, through center-coincidence and equal-radius constraints. After the parameter edit, these sketch circles update together with their referenced geometry (A4 and A6), preserving the intended geometric relationships. Together, these observations provide qualitative evidence that DIP-CAD preserves design intent through explicit geometry references, accurate sketch placement, and persistent external constraints.

\subsection{Ablation Studies}
\label{subsec:ablation}

We evaluate six ablations on GERD-EVL under the same static and editing protocols. The first three remove mechanisms for geometry reference, sketch coordinate bridge, and sketch external constraints, respectively. The remaining ablations examine the supporting roles of registration, retrieval, and execution verification.

\begin{enumerate}
    \item \textbf{w/o EGR (explicit geometry reference):} Stage~1 removes edge ID tags from $I_t^{\mathrm{tag}}$, and removes $T_t^{\mathrm{geom}}$. CadQuery selectors replace explicit entity IDs.
    \item \textbf{w/o LCB (local coordinate bridge):} Stage~2 removes the bridge from view coordinates to sketch local coordinates. The VLM must predict sketch primitive coordinates in the sketch plane's local coordinate system directly.
    \item \textbf{w/o SEC (sketch external constraints):} Stage~2 disables sketch constraints between sketch primitives and projected external geometry.
    \item \textbf{w/o Reg (registration):} Stage~1 disables image registration and the associated pseudo-bounding sphere $S_t$. Stage~2 removes the $[0,1]$ restriction on predicted sketch coordinates.
    \item \textbf{w/o RAG (retrieval augmented generation):} Stage~1 does not retrieve similar modeling examples from GERD-RET. The VLM predicts the next command using only the current model and target image.
    \item \textbf{w/o EVf (execution verification):} Stage~3 (preview-based verification) is removed. Results generated from Stages~1 and~2 are committed without visual verification.
\end{enumerate}

\begin{figure}[pos=htbp]
    \centering
    \includegraphics[width=1.0\linewidth]{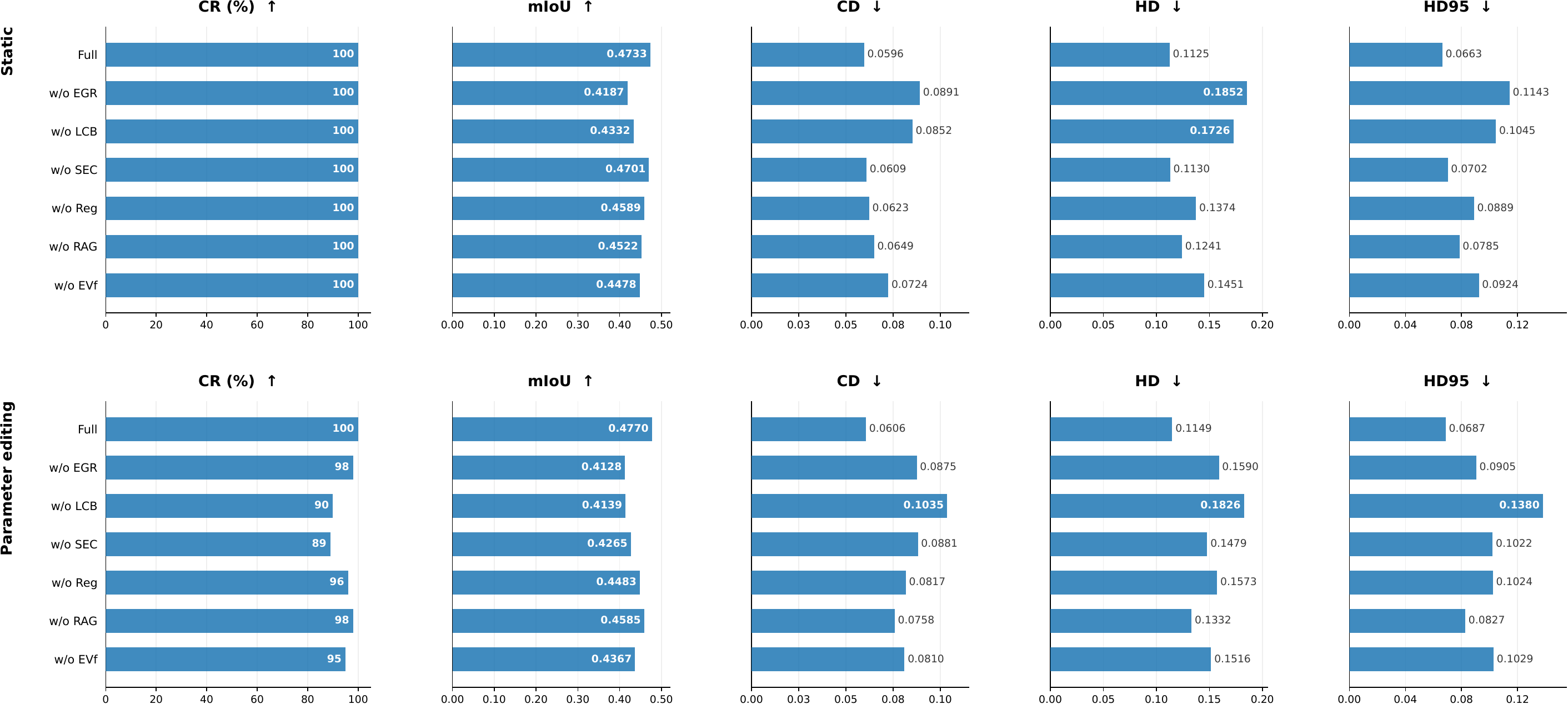}
    \caption{\textbf{Results of ablation studies.}}
    \label{fig:ablation}
\end{figure}

All configurations achieve $100.0\%$ CR in static reconstruction, but the full system gives the best geometric scores. Editing exposes larger differences in execution success (\Cref{fig:ablation}).

Removing explicit geometry reference (EGR) causes the largest static mIoU drop, from $0.4733$ to $0.4187$. After editing, CR falls to $98.0\%$ and mIoU to $0.4128$, the lowest among the ablations. This supports the importance of accurate geometry selection.
Without the local-coordinate bridge (LCB), static mIoU falls to $0.4332$ and editing CR to $90.0\%$. 
% This ablation isolates the conversion from view coordinates to native sketch coordinates within the broader placement process, which also uses edge labels and projected samples. 
The results support the role of deterministic coordinate conversion in generating parametric sketches.
% whose geometry and subsequent updates are consistent with their selected supports.
Removing sketch external constraints (SEC) has a modest static effect: mIoU falls only to $0.4701$. After editing, however, CR falls to $89.0\%$ and mIoU to $0.4265$. 
% This ablation removes both projected edge samples and automatic constraints to external geometry, examining their combined contribution to placement and dependency construction. 
Together with the case study traces, the larger differences after editing support the importance of encoding relationships to source geometry as persistent sketch constraints.

Registration, RAG, and execution verification also improve geometric scores. Removing them reduces editing CR to $96.0\%$, $98.0\%$, and $95.0\%$, respectively. Their contributions complement the reference mechanisms through visual alignment, modeling examples, and rejection of incorrect operations.

\section{Conclusions}
\label{sec:conclusion}

We presented \methodname, a design-intent-preserving agent harness that integrates VLM reasoning with CAD-kernel execution for single-image parametric CAD modeling. It follows a stepwise workflow of command generation, sketch generation, and modeling verification. For modeling commands, edge-ID tags and geometry tables enable explicit geometry reference. For parametric sketches, a view-to-local coordinate bridge ensures accurate sketch placement, while automatic constraint generation establishes persistent dependencies on projected external geometry. Together, these mechanisms enable the generation of native parametric CAD models with editable sketches, feature histories, and dependencies that support parameter-driven updates. We also introduced GERD, a dataset of 2,700 parts that aligns modeling commands, intermediate models, and geometry IDs to support research on parametric CAD modeling.

On GERD-EVL and a DeepCAD test subset, \methodname\ achieves the highest geometric accuracy among the evaluated methods, with a 100\% compilation success rate in both static reconstruction and parameter editing evaluations. It improves static reconstruction mIoU by 11.1\% and 5.6\%, respectively. After increasing all extrusion depths by 10\%, the corresponding mIoU improvements are 7.0\% and 4.2\%. Parameter editing case studies demonstrate how explicit dependencies enable downstream geometry to update with parameter edits. Ablation studies further confirm the importance of the proposed mechanisms, particularly for preserving editability. These results highlight that accurate static geometry alone is insufficient to ensure the preservation of design intent.

\methodname\ has limitations arising from its use of a single rendering viewpoint and dense edge-ID annotations on complex parts. It also does not yet support references to modeling commands or sketch primitives. Future work will explore multiple views, active viewpoint planning, adaptive entity visualization, and richer reference types to improve modeling flexibility and scalability.

% -------------------- Declarations --------------------
\section*{CRediT authorship contribution statement}
First Author: Conceptualization, Methodology, Software, Validation,
Writing -- original draft. Second Author: Supervision, Writing -- review and editing.

\section*{Declaration of competing interest}
The authors declare that they have no known competing financial interests or
personal relationships that could have appeared to influence the work reported
in this paper.

\section*{Funding}
This research did not receive any specific grant from funding agencies in the
public, commercial, or not-for-profit sectors.

\section*{Data availability}
The data and code supporting the findings of this study will be made available
as requested.

% Add this section only when generative AI or AI-assisted technologies were
% used in manuscript preparation. Adapt the statement to the actual use.
% \section*{Declaration of generative AI and AI-assisted technologies in the manuscript preparation process}
% During the preparation of this work, the authors used [TOOL/SERVICE] to
% [PURPOSE]. The authors reviewed and edited the output and take full
% responsibility for the content of the article.

% -------------------- Appendices --------------------
\appendix
\section{Supplementary Material}
\label{app:supplementary}

% 本材料用于补充正文中未提及的细节内容，具体包含GERD数据集，检索数据集、以及用于Stage3审核的建模预览图三部分。

\subsection{Implementation Details}

The implementation follows the three-stage workflow described in the main paper. Stage~1 takes the target image and the current CAD state as input, performs view registration, and retrieves relevant modeling examples from GERD-RET to provide geometric and operational references. Stage~2 is conditionally activated when the planned operation requires a new sketch. In this stage, the VLM generates sketch geometry based on the selected support plane and projected external references. Stage~3 generates a visual preview of the candidate operation and evaluates its consistency with the target. The verification result includes an ACCEPT/REJECT verdict, an explanation, and a confidence score.

All stages of \methodname\ use gpt-5.6-sol as the VLM, with at most two failed verification attempts in Stage~3. Stage~1 retrieves up to three examples with $\lambda=0.6$. We use FreeCAD 1.1 and render CAD viewport images at $500 \times 500$ pixels. Generation terminates when END is predicted or after 25 modeling steps.

For image-based retrieval, we use DINOv3 CLS features and Euclidean distance without additional task-specific fine-tuning. The retrieved examples provide both visual references and operation-type information to support subsequent command generation.

Training-based baselines are trained on the GERD-EVL training split to generate CADFS code~\citep{pyatov2026cadfs}. Other baselines generate FreeCAD or CadQuery code and use gpt-5.6-sol unless specified. Evaluation on DeepCAD uses the 100-part CADCoder-test100 subset because evaluating the full test set is computationally costly~\citep{doris2026cadcoder}.

% All stages of \methodname\ use gpt-5.6-sol as the VLM, with at most two failed verification attempts in Stage~3. Stage~1 retrieves up to three examples with $\lambda=0.6$. We use FreeCAD 1.1 and render images at $500 \times 500$ pixels. Generation terminates when END is predicted or after 25 modeling steps.

% Training-based baselines are trained on the GERD-EVL training split to generate CADFS code~\citep{pyatov2026cadfs}. Other baselines generate FreeCAD or CadQuery code and use gpt-5.6-sol unless specified. Evaluation on DeepCAD uses the 100-part CADCoder-test100 subset because evaluating the full test set is computationally costly~\citep{doris2026cadcoder}.

\subsection{GERD data structure}

\begin{figure}[pos=t]
    \centering
    \includegraphics[width=0.75\linewidth]{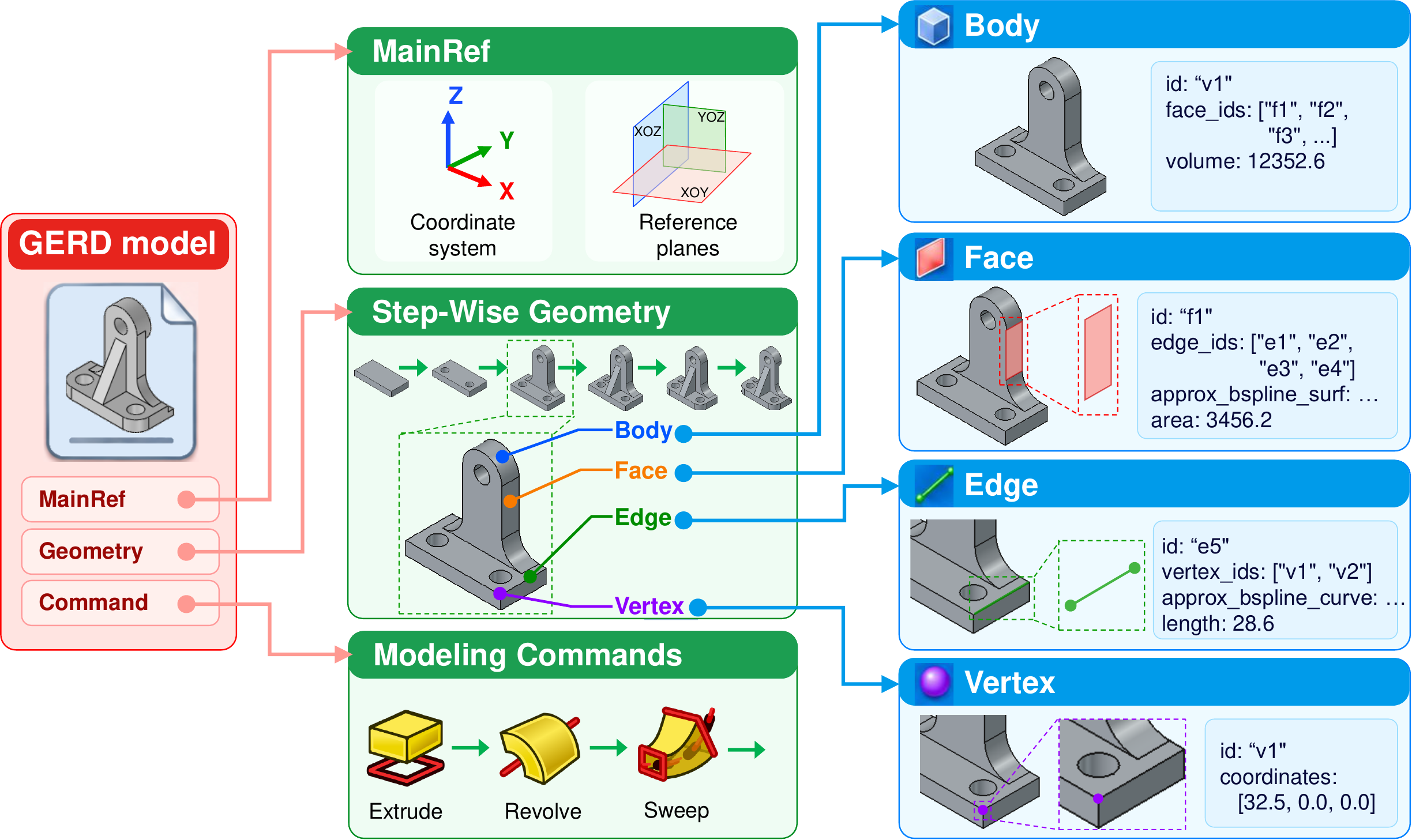}
    \caption{\textbf{GERD data structure.} Each model contains MainRef, Geometry, and Command. MainRef stores three main reference planes and their coordinate systems: $Z$ is normal to the plane, and $X/Y$ define local coordinates. Geometry indexes four levels by ID. Bodies store boundary face IDs and volume. Faces store surface type, boundary edge IDs, and area. Edges store curve type, endpoint vertex IDs, and length. Vertices store 3D coordinates. Faces and edges also provide approximate B-spline representations for reconstruction across CAD systems. MainRef, Geometry, and Command share the same ID namespace.}
    \label{fig:gerd}
\end{figure}

GERD is designed to preserve the associations between parametric modeling operations and the evolving geometric structure of a CAD model. As illustrated in \Cref{fig:gerd}, each model consists of three interconnected components: MainRef, Geometry, and Command. MainRef defines the primary reference planes and their local coordinate systems, providing a consistent spatial reference for sketch placement and feature construction. Each reference plane has an associated local coordinate system, where the $Z$-axis represents the plane normal and the $X/Y$-axes define its local coordinates.

Geometry organizes the boundary representation (B-Rep) into a four-level hierarchy: Body, Face, Edge, and Vertex. Each Body records its constituent face IDs and volume. Each Face stores its surface type, boundary edge IDs, and area. Each Edge records its curve type, endpoint vertex IDs, and length, while each Vertex stores its three-dimensional coordinates. In addition to these topological and geometric attributes, Faces and Edges provide approximate B-spline representations of their underlying surfaces and curves, respectively, facilitating geometric reconstruction across different CAD systems.

The Command component records the parametric modeling operations and their associated information. Importantly, MainRef, Geometry, and Command share a common ID namespace, allowing modeling operations to explicitly reference geometric entities and reference elements. This design preserves the correspondence between a modeling command, its geometric effects, and the entities involved in the operation, rather than representing a CAD model as an isolated final shape.

Beyond its use in GERD-RET and GERD-EVL, this structured representation supports the derivation of datasets with different viewpoints, rendering modalities, and supervision targets. The explicit entity references enable tasks such as next-command prediction, geometric entity selection, B-Rep representation learning, and geometric reference validation. By unifying modeling commands with step-level geometry and entity references, GERD provides a reusable data foundation for studying parametric dependencies and design-intent-preserving CAD modeling.

% Beyond applications of GERD-RET and GERD-EVL, GERD supports the derivation of datasets with different viewpoints, rendering modalities, and supervision targets. Its explicit entity references enable tasks such as next-command prediction, geometric entity selection, B-Rep representation learning, and geometric reference validation. By unifying modeling commands with step-level geometry and entity references, GERD provides a reusable data foundation for studying parametric dependencies and design-intent-preserving CAD modeling.

\subsection{Retrieval dataset}

\begin{figure}[pos=htbp]
    \centering
    \includegraphics[width=0.6\linewidth]{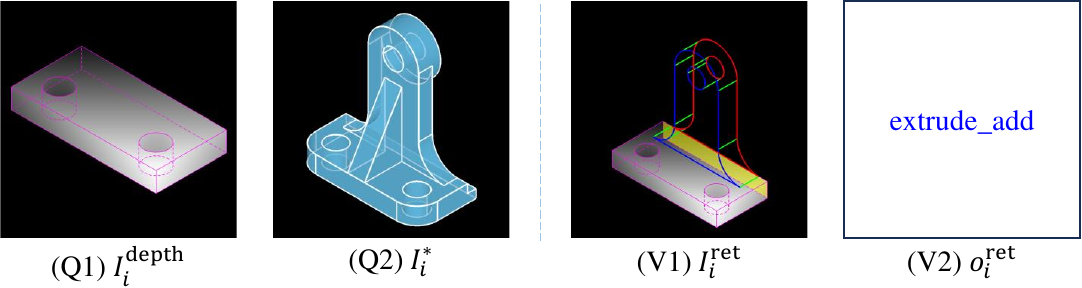}
    % 总标题
    \caption{\textbf{An example retrieval instance.} Each instance is $\{I_i^{\mathrm{depth}},I_i^*;I_i^{\mathrm{ret}},o_i^{\mathrm{ret}}\}$, where $i$ is the instance index. (Q1), (Q2): the retrieval queries, which contain the current depth image and completed-part image. (V1), (V2): the retrieved results, which contain a pseudo-preview and its command type. In (V1), yellow highlights the sketch support, red and blue outlines mark the starting and ending profiles, and green segments connect corresponding profile points. Existing geometry retains depth shading and magenta edge outlines.}
    \label{fig:ret-data}
\end{figure}

GERD-RET is organized around intermediate modeling states rather than isolated final CAD models. Each retrieval instance consists of a query pair and its corresponding retrieval result, as illustrated in \Cref{fig:ret-data}. Specifically, the query contains the depth image of the current partially constructed model, $I_i^{\mathrm{depth}}$, and the image of the completed target model, $I_i^*$. These two views provide complementary information: the former describes the existing geometry, while the latter specifies the desired final shape. Their combination allows the retrieval process to consider both the current modeling state and the geometric changes required to reach the target.

The retrieved result contains a pseudo-preview image, $I_i^{\mathrm{ret}}$, and its associated operation type, $o_i^{\mathrm{ret}}$. Rather than returning an operation label alone, the pseudo-preview provides geometric information about how a candidate operation relates to the existing model. As shown in \Cref{fig:ret-data}, the sketch support is highlighted in yellow, while red and blue outlines indicate the starting and ending profiles, respectively. Green line segments connect corresponding profile points to illustrate their geometric correspondence. The existing geometry retains depth shading and magenta edge outlines, making the relationship between the candidate feature and the current model visually distinguishable.

For image-based matching, DINOv3 CLS features are extracted and compared using Euclidean distance without task-specific fine-tuning. During inference, Stage~1 retrieves up to three relevant examples using $\lambda=0.6$. The retrieved pseudo-previews and operation types serve as additional visual and semantic references for the VLM, helping it infer plausible modeling operations and their geometric configurations.

\subsection{Registration}

View registration is essential for maintaining consistent geometric references between the CAD viewport and the target image. Since the system relies on visual observations to identify support faces and construct sketches, changes in viewport framing can affect the visibility and spatial correspondence of existing geometric entities. To reduce unnecessary visual variations, camera registration is restricted to panning and zooming, while the viewing direction is controlled separately according to the selected sketch plane.

A particular challenge arises when the current model occupies most of the viewport. Directly applying \texttt{FitAll()} to the existing geometry may enlarge the model excessively, leaving insufficient space for the sketch profile of the next feature. To address this issue, we introduce a pseudo-bounding sphere $S_t$, constructed from the axis-aligned bounding box (AABB) of the current model. Its center is aligned with the center of the AABB, and its radius is selected to enclose the bounding box. The sphere is included as a temporary reference during viewport fitting, allowing \texttt{FitAll()} to account for an extended spatial region without modifying the actual CAD geometry or feature history.

\begin{figure}[pos=htbp]
    \centering
    \includegraphics[width=0.6\linewidth]{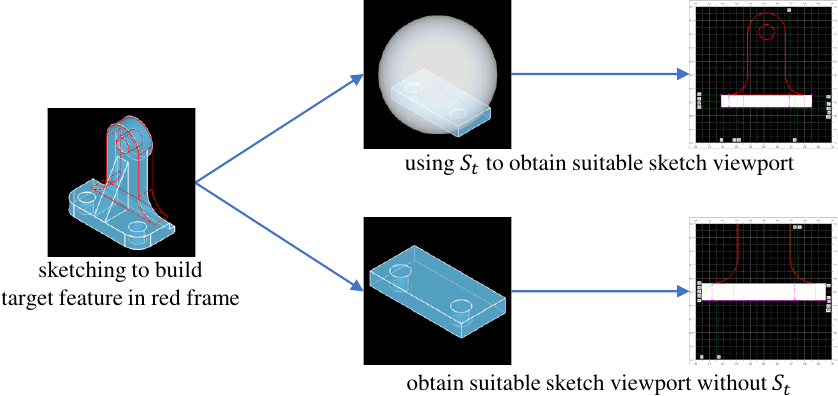}
    % 总标题
    \caption{\textbf{Effect of the pseudo-bounding sphere $S_t$ on sketch viewport framing.} The back plate highlighted by the red frame is constructed through sketching and extrude. Including the pseudo-bounding sphere $S_t$ in \texttt{FitAll()} leaves space for the full sketch profile. But fitting only the existing base enlarges it within the viewport and clips the upper part of the intended profile.}
    \label{fig:register}
\end{figure}

As demonstrated in \Cref{fig:register}, including $S_t$ preserves additional viewport space for the intended sketch profile, whereas fitting only the existing base may cause the upper portion of the profile to be clipped. After viewport framing, an orthographic view is established along the normal of the selected support face. The sketch plane is then mapped to a normalized $[0,1] \times [0,1]$ coordinate frame with scale markings. Existing boundary geometry is projected onto this plane as external references, with approximately 6--8 sampling points per edge in addition to its endpoints. These projected references provide explicit geometric context for subsequent sketch generation.

\subsection{Modeling Preview}

Modeling previews provide an explicit visual representation of the geometric changes introduced by a candidate operation. Instead of relying solely on the final rendered model, the system highlights the expected geometric modification and its associated operation direction. This representation makes different types of modeling operations visually distinguishable and provides additional evidence for Stage~3 verification.

Different colors are assigned to different geometric elements according to their modeling semantics. Additive operations, such as extrusion, are displayed using green geometry, while subtractive operations are represented in red. Red arrows indicate the operation direction. For rotational and sweeping operations, the corresponding rotation axis and sweep path are highlighted in cyan. Fillet and chamfer operations highlight the affected edges in blue, while the END state is represented in gray. These visual cues allow the VLM to distinguish the intended geometric modification from the existing model and interpret the spatial configuration of the candidate operation.

\begin{figure}[pos=htbp]
    \centering
    % 左子图
    \begin{subfigure}{0.15\linewidth}
        \centering
        \includegraphics[width=\linewidth]{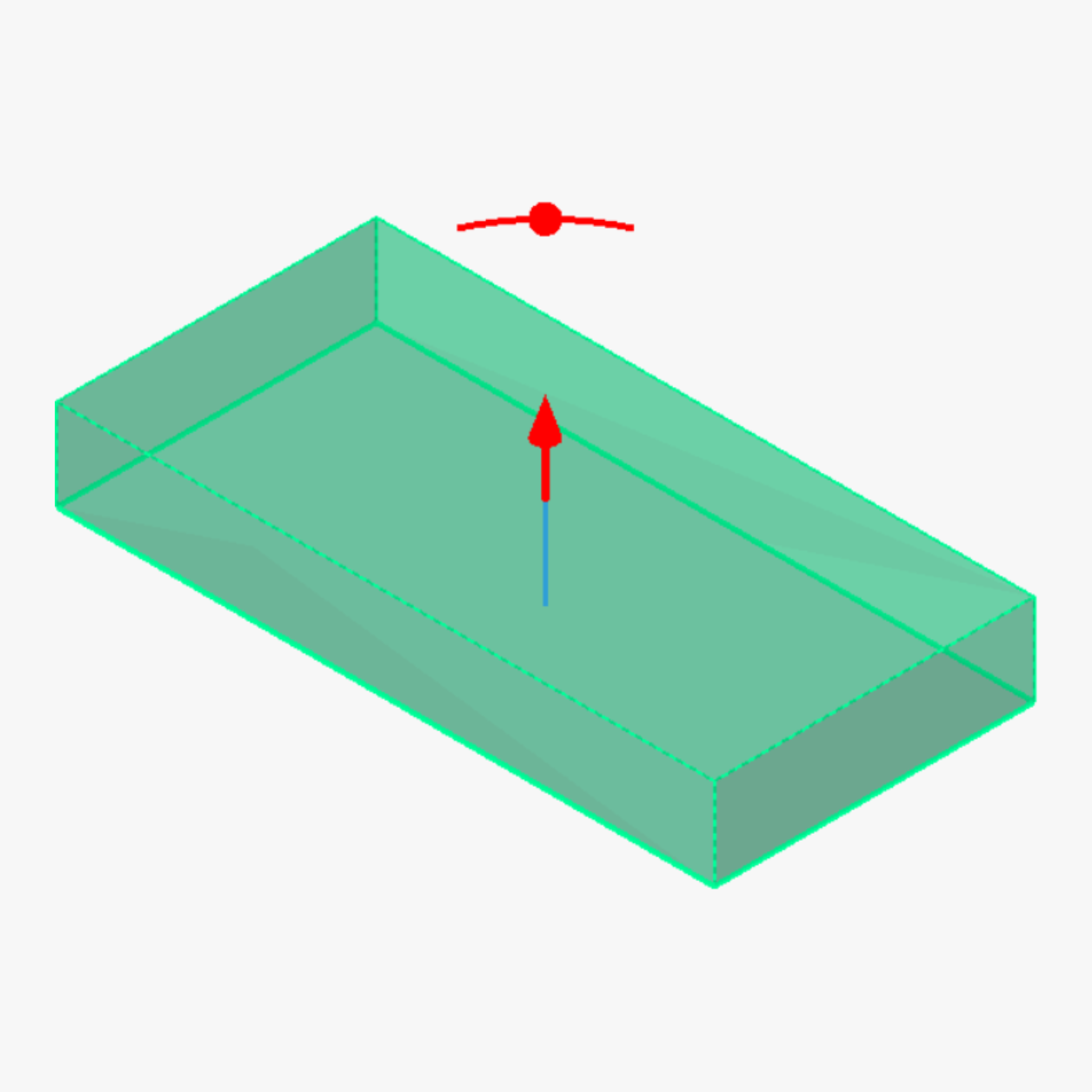}
    \end{subfigure}
    % 右子图
    \begin{subfigure}{0.15\linewidth}
        \centering
        \includegraphics[width=\linewidth]{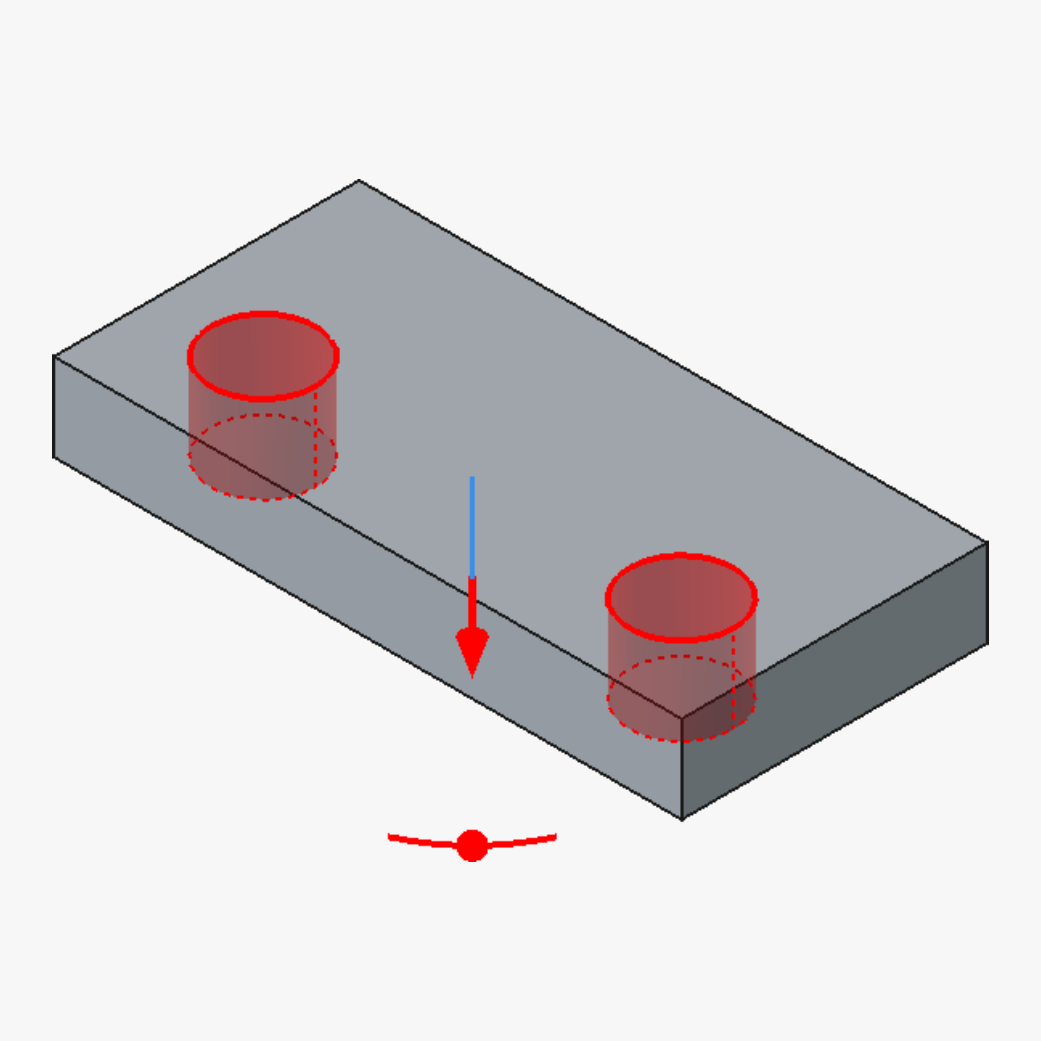}
    \end{subfigure}
    \begin{subfigure}{0.15\linewidth}
        \centering
        \includegraphics[width=\linewidth]{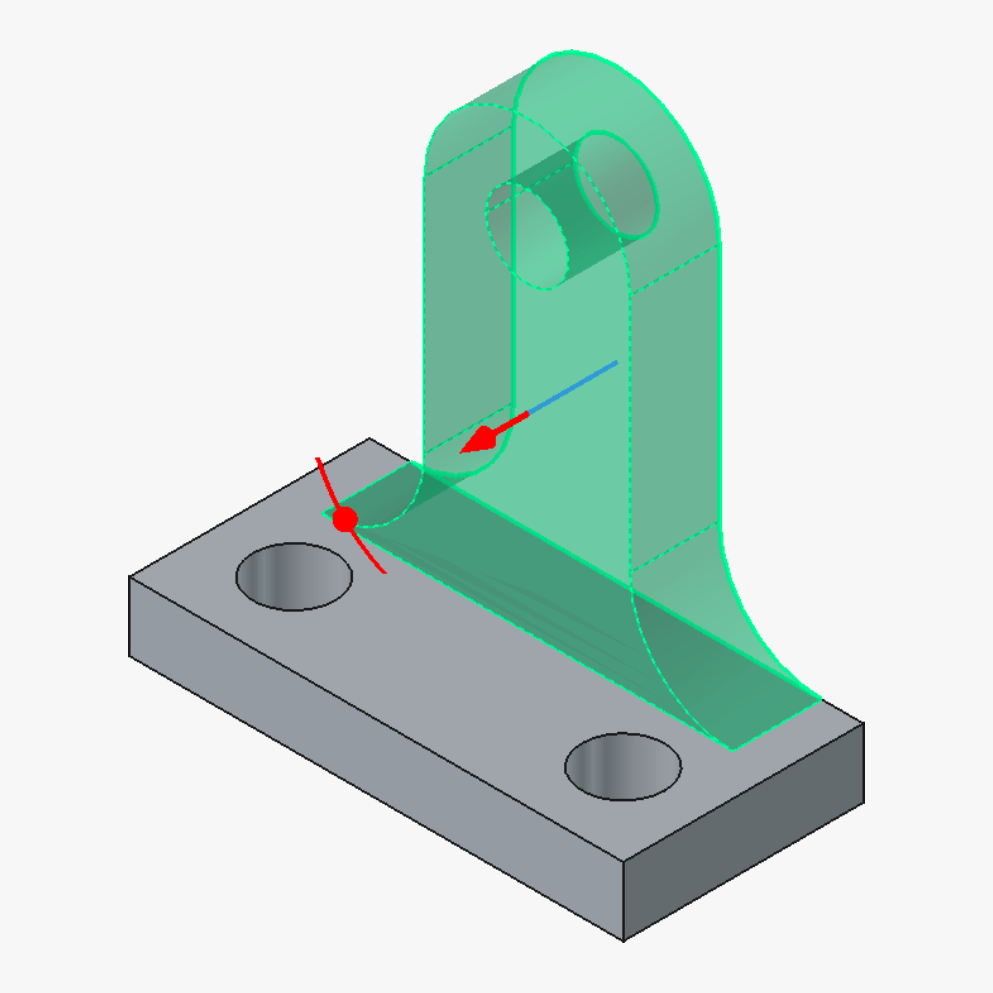}
    \end{subfigure}
    \begin{subfigure}{0.15\linewidth}
        \centering
        \includegraphics[width=\linewidth]{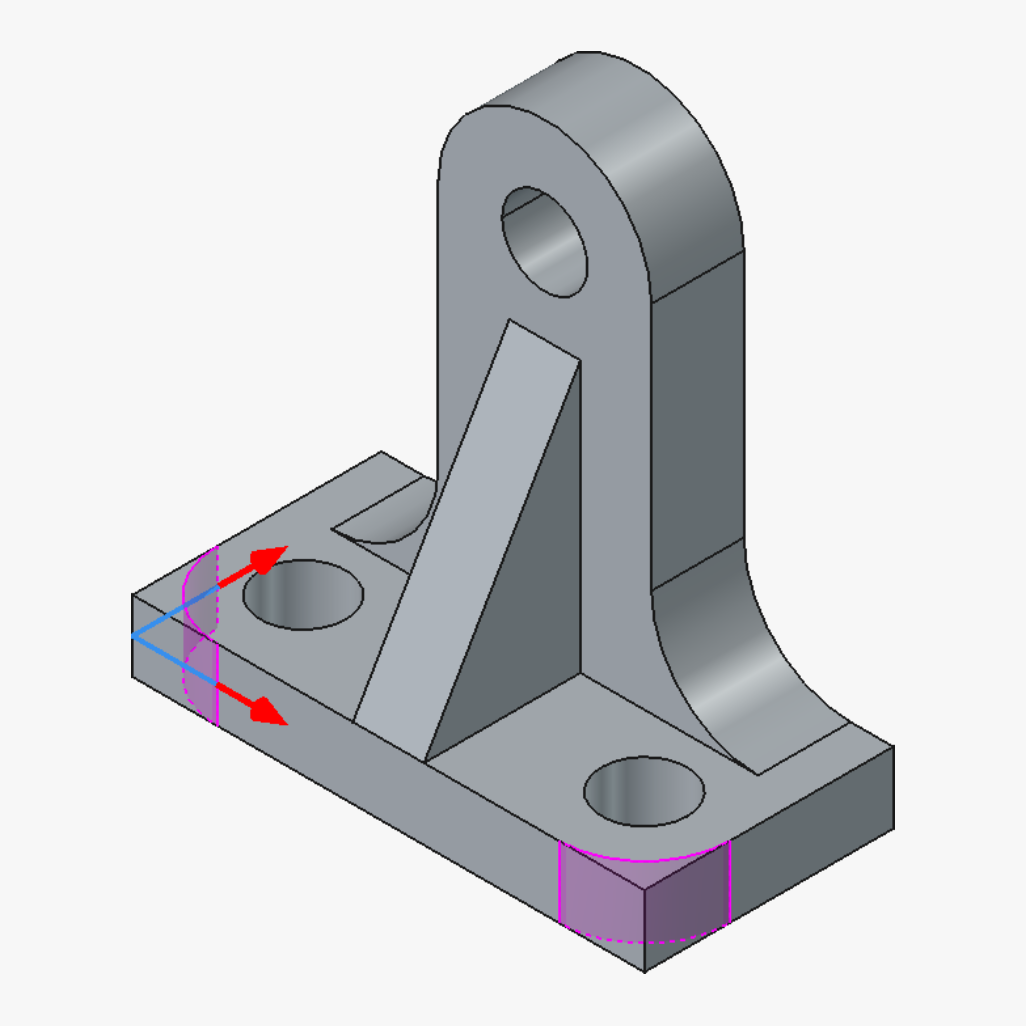}
    \end{subfigure}
    % 总标题
    \caption{\textbf{Modeling previews for verification.} From left to right: adding the base, cutting two holes, adding the back plate, and rounding the base corners. The previews expose the geometric change and operation direction, so that the VLM can compare the candidate with the target.}
    \label{fig:fcad_preview}
\end{figure}

\Cref{fig:fcad_preview} illustrates four representative modeling previews in a sequential construction process: adding the base, cutting two holes, adding the back plate, and rounding the base corners. Each preview exposes the geometric effect of the corresponding operation, including the modified regions and operation direction. The resulting preview is provided to the VLM together with the current CAD state and target image for visual verification. The VLM then determines whether the candidate operation is consistent with the target and returns an ACCEPT/REJECT verdict, an explanation, and a confidence score. If the candidate is rejected, the feedback is used to guide subsequent correction attempts, with at most two failed verification attempts allowed in Stage~3.

% Add supplementary derivations, implementation details, or results here.

\bibliographystyle{cas-model2-names}
\bibliography{references}

\end{document}